\documentclass{article}
\usepackage[english]{babel}
\usepackage[utf8]{inputenc}
\usepackage{johd}
\usepackage{subfig}
\usepackage{amsfonts}
\usepackage{amsmath}

\title{Early Warning Signals of Social Instabilities in Twitter Data}

\author{Vahid Shamsaddini$^{b}$, Henry Kirveslahti$^{a}$, Raphael Reinauer$^{a}$, \\
Wallyson Lemes de Oliveira$^{b}$$^{*}$, Matteo Caorsi$^{b}$ , Étienne Voutaz$^{c}$\\
\\
        \small $^{a}$EPFL, Route Cantonale, 1015, Lausanne, CH \\
        \small $^{b}$L2F SA, Rue du Centre 9, 1025, Saint-Sulpice, CH \\
        \small $^{c}$ Armasuisse, Feuerwerkerstrasse 39, 3602, Thun, CH \\
        \small $^{*}$Corresponding author: Wallyson Lemes de Oliveira; \tt{w.lemes@giotto.ai}
}

\date{} 

\begin{document}

\maketitle

\begin{abstract} 
\noindent The goal of this project is to create and study novel techniques to identify early warning signals for socially disruptive events, like riots, wars, or revolutions using only publicly available data on social media. Such techniques need to be robust enough to work on real-time data: to achieve this goal we propose a topological approach together with more standard BERT models. Indeed, topology-based algorithms, being provably stable against deformations and noise, seem to work well in low-data regimes. The general idea is to build a binary classifier that predicts if a given tweet is related to a disruptive event or not. The results indicate that the persistent-gradient approach is stable and even more performant than deep-learning-based anomaly detection algorithms. We also benchmark the generalisability of the methodology against out-of-samples tasks, with very promising results.  \end{abstract}

\noindent\\

\tableofcontents

\section{Context and motivation}

The problem concerns the individuation of early warning signals for the outburst of revolutions, riots, or wars. It has been proven in numerous studies -- \citealp{lim2011tahrir, tufekci2012social, lim2012clicks, lotan2011arab, voutaz2022early} -- that it is possible to extract early warning signals of contemporary activism from social media. Examples of such events are the Arab Spring of the Tahrir square in 2011 and farmers’ protests in India and Iraq of the years 2019 and 2020. Social media such as \url{https://twitter.com/?lang=en} are becoming more and more present in everydays life and are often used by people as a mean to express their thoughts or concerns. Hence, tweets likely contain valuable information  to identify early warning signals for disruptive events. In this project, we will try to create and study novel techniques to identify early warning signals on tweet data.

\section{Dataset description and preprocessing}

To identify the events of importance we used ACLED website -- \url{https://acleddata.com/dashboard/#/dashboard} -- where riots, protests, wars, and other kinds of social revolutions are monitored daily all over the world. 

The following are the events that we identified and on which we will perform our analysis:

\begin{table}[H]
\centering 
\caption{\label{tab1} Selected protests}
\begin{tabular}{lp{2.3cm}llp{3cm}p{1.8cm}}
\hline
type & name & cause & location & date & organisers' estimate \\
\hline

protest	& Put it to the People march & Second EU referendum & London	& 23 March 2019	& 1,200,000 \\
protest	& Let Us Be Heard	& Second EU referendum &	London & 19 October 2019 &	1,000,000 \\
protest-riot & George Floyd Protests & Black Lives Matter & USA  & May 26 – June 20, 2020 & 15,000,000 - 26,000,000 \\
protest & 2017 Women's March & Feminism	 & USA &	January 21, 2017 & 3,300,000 – 5,600,000 \\
protest	 & March for Our Lives	 & Gun control & 	USA  & 	March 24, 2018 & 	1,200,000 - 2,000,000\\
protest	 & 2018 Women's March	 & Feminism	 & USA 	 & January 20, 2018	 & 1,500,000 \\
protest & 	\#RickyRenuncia & 	Anti-corruption & 	San Juan & 	July 8, 2019 & 	1,100,000 \\
protest	 & 2012 Quebec student protests  & 	raise university tuition & 	Quebec	 & February 13, 2012 – September 7, 2012	& - \\ 

\hline
\end{tabular}
\end{table}

These events have been identified based on a few criteria:

\begin{itemize}
    \item[1.] Only English-speaking countries have been selected: this simplifies the use of pre-trained models
    \item[2.] Events with high mediatic resonance have been selected
\end{itemize}
\subsection{Fetching the data}
To get the data Twitter API was used. We fetched data monthly, and for each month, we fetched up to 500 tweets according to API limitations. In the table below we show an example of retrieved tweets:

\begin{table}[H]
\centering 
\caption{\label{tab2} Tweet examples}
\begin{tabular}{p{8cm}c}
\hline
tweet & id \\
\hline

\texttt{\#WomensMarchOnCanberra \#March4Justice \#crediblewomen \#enough} !!!  >:( \texttt{\#auspol\#WomensMarch} 15th March 2021
Check for updates of marches in your city at 
\texttt{@womensmarchaus} \url{https://t.co/JBnGZ3Y3aF}
& 1366932844539383808 \\
Our movement is growing and now we need your help please sign our petition we will take this to Parliament  on March 15 \texttt{\#March4Justice} \url{https://t.co/bHat9xnbdR}
& 1366932764893712384 \\
Who else is just insanely angry?
\texttt{\#March4Justice 
\#auspol}	& 1366903350369787906 \\

\hline
\end{tabular}
\end{table}

Twitter API allows us to fetch tweets that contain a specific class of hashtags and keyboards. To fetch tweets related to each event, we use the following class of hashtags and keyboards. 

\begin{table}[H]
\centering 
\caption{\label{tab-hashtags}Hashtags and Keywords}
\begin{tabular}{|p{3cm}||p{7cm}|p{3cm}|} 
\hline
Event & Hashtags & Keywords \\
\hline

London Protests	& \texttt{\#peoplesvote, \#Brexit} & Brexit \\
 \hline
George Floyd Protests & \texttt{\#BlackLivesMatter \#GeorgeFloyd \#GeorgeFloyd}	&  \\
 \hline
Feminism related protests & \texttt{\#WomensMarch, \#WMW, \#WhyWeMarch} & Feminism	  \\
 \hline
Gun control & \texttt{\#MarchForOurLives, \#NeverAgain, \#GunControlNow, \#EnoughIsEnough} & \\
\hline
\end{tabular}
\end{table}

\subsection{Preprocessing}
In order to improve the performance of the model, we applied some standard preprocessing techniques such as
\begin{itemize}
\item Hashtag removal: we remove hashtags to prevent the model from overfitting the hashtags
\item Html tags removal: we remove all URLs and HTML tags that could not have valuable information
\item Emoji removal: we remove all emojis and punctuation marks
\item Lower case only: we change all letters to lowercase to improve the sentence embedding.
\end{itemize}

\subsection{Data labelling}
We have framed our problem into a binary classifier, where the tweets are classified as event-related or non-event related. We will now present how we constructed our binary dataset.

To understand if a tweet is related to a specific event or not, we searched for hashtags that people have used to write tweets specific to that event. Once we have found the proper hashtags for each event, we labeled tweets containing them as event-related tweets (class 1). On the other hand, the class of non-event related tweets (class 0) consists of random tweets on subjects like sports, science, fiction, etc. Subsequently, we have an extensive collection of tweets with an associated label that is either 1 (event-related) or 0 (non-event related).

\begin{figure}[H]
\centering
\includegraphics[width=0.8\textwidth]{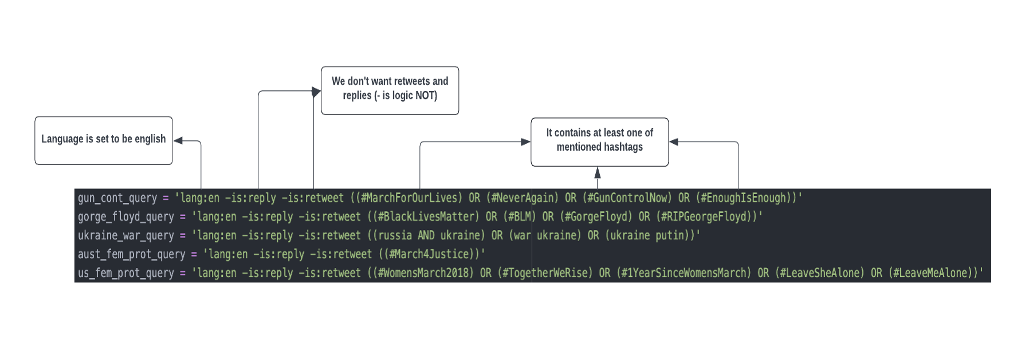}
\caption{\label{fig1}Example of a twitter query.}
\end{figure}

To get the non-event-related tweets (class 0), the NOT operator was applied to the queries in Figure \ref{fig1}.

Importantly, the hashtags were removed from the tweets to avoid model overfitting. This is to ensure that the model will learn from the entire sentence of the tweet and not only the hashtag.


\section{Methods}
We benchmark multiple methods in this study, and analyze the performance of each in different data abundance regimes.

\subsection{Persistence gradient model}
\label{subsec:pg}

\begin{figure}[H]
\centering
\includegraphics[width=0.8\textwidth]{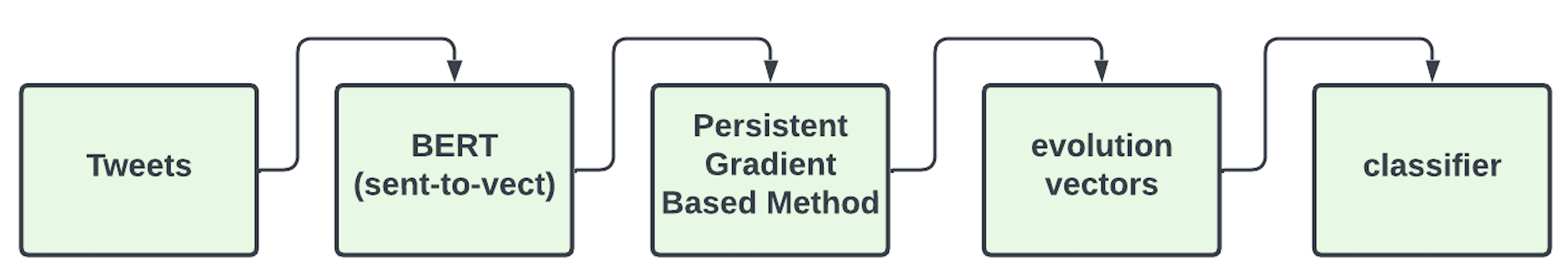}
\caption{\label{fig_steps_of_app}Different steps of our approach.}
\end{figure}

Persistent gradient techniques are described in detail in section \ref{sec:pg1}. Figure \ref{fig_steps_of_app} shows how persistent gradients are used for the classification. Concisely, once we get tweets as input, we use a sentence-to-vector model to embed tweets as points in Euclidean space, thus obtaining a point cloud $X$ in Euclidean space (see \ref{sec:pg2} for more details). Afterwards, we apply the persistent gradient method to this point cloud, and we modify the points following the gradient directions for $30$ cycles. The variable $X'$ will be the difference vector between the points' locations before and after the gradient evolution: $X'$ is the input of a classifier. This classifier is tasked to predict if a tweet is related to a protest or not. In our analysis we will use three classifiers:
\begin{enumerate}
    \item Logistic Classifier (LR): In this classifier, our probabilities $Y$ are modeled as  the sigmoid of $\beta X'$.  
    
    \item Random Forest Classifier (RF): Here, $Y$ is determined as the majority vote of an ensemble of decision trees obtained by fitting on samples of the training set and the covariates $X'$.
    \item Gradient Boost Classifier (GB):  In this model, $Y$ is determined as the result of an optimization algorithm that combines weak learners in a recursive fashion. Each learner is tasked to reduce the error of the previous model by fitting on its residuals.
\end{enumerate}

\subsection{Vanilla gradient model}

Let $X \in \mathbb{R}^{n \times d}$ be the point cloud of $n$ embedded tweets in $d$ dimensions. We use standard gradient methods on the loss function 
$$||X-x_0||_2^2,$$
where $x_0$ is the barycenter. In a subsequent step, we push the points with the same procedure described in section \ref{subsec:pg} and use the difference vector as our new feature $X'$. The classifiers trained on this feature are the same as in section \ref{subsec:pg}.

\subsection{BERT}

As a benchmark, we use a pre-trained Bidirectional Encoder Representation from Transformers (BERT), widely considered as the state of the art in natural language processing. The particular model we consider was presented in \citealp{DBLP:journals/corr/abs-1810-04805} and is called the  \texttt{bert-base-uncased} from HuggingFace\footnote{available on \url{https://huggingface.co/bert-base-uncased}}. This is a Transformer model with $L=12$ transformer blocks, hidden size $H=768$ and $A=12$ attention heads.

To preprocess the tweets, we follow the same procedure as with the other methods described in Section 2.1.
However, instead of using a \texttt{sent2vec} we use BERT. This method serves essentially the same purpose as the \texttt{sent2vec}, but it makes the tweets amenable to the multi-head attention structure, which allows to produce contextualized embeddings. In the theory, this should solve the problem of mapping multiple-meaning words into the same vector.


We fine-tune the model with 4 epochs using the Adam optimizer with learning rate $2 \times 10^{-5}$ and $\epsilon=10^{-10}$.


\subsection{Feed-forward model}

Additionally, we benchmarked against a standard feed-forward network with \texttt{reLU} activation functions with 3 layers containing respectively 768, 100, and 2 nodes. We fitted on our point cloud $X$ after representing our tweets with sentence embedding (see \ref{sec:pg2}). This network is fully connected and is tasked to predict the class of tweets.


\section{Results and discussion}
In this section, we show the results of our benchmark. We have performed in-sample and out-of-sample analysis, i.e. using validation data from the same or different events respectively.

The correlation analysis is also useful to see how similar the different method's predictions are.

\subsection{Correlation}
\label{sec:corr}
\begin{figure}[H]
\label{fig:corr}
\centering
\includegraphics[width=0.8\textwidth]{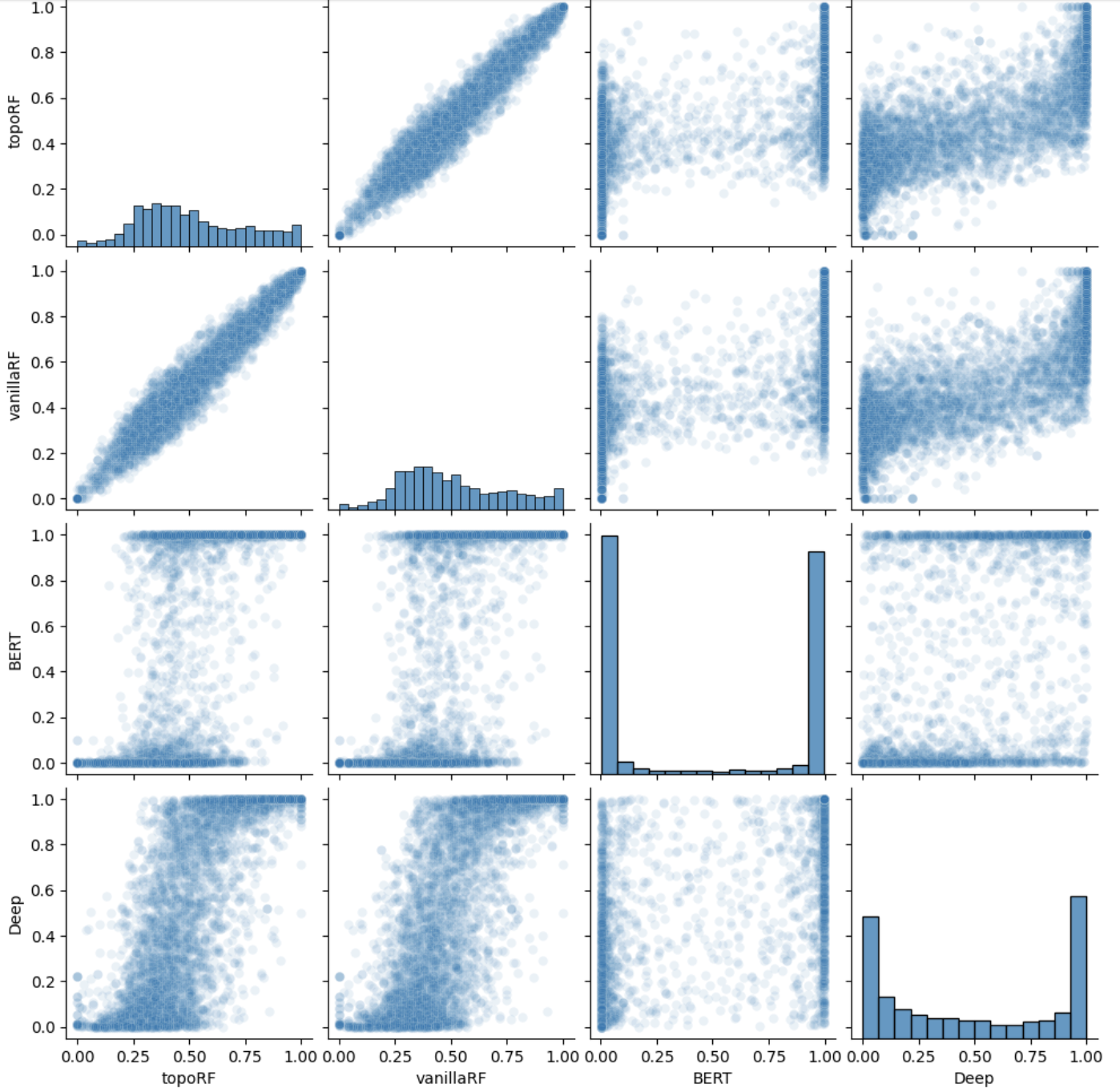}
\caption{\label{fig7}Correlation plot between Topological gradient Random Forest (\textbf{topoRF}), Vanilla gradient Random Forest (\textbf{vanillaRF}), \textbf{BERT} and Feed-forward network(\textbf{Deep})}
\end{figure}

First, let us assess how correlated the different models are. In Figure \ref{fig7}, we compare the predictions of the different methods. To perform this analysis, we first stored the probabilities of class one for each model (per sample). Data from the \texttt{George Floyd Protest} was used to train and test all models.

We see that the persistence gradient with RF and the vanilla gradient with RF are very correlated.(see \ref{subsection:Vanilla-Pers} for more details) BERT does not seem to correlate much with any of the other models.
These results suggest using simple ensembling techniques (max ensembling, mean ensembling) to merge the effectiveness and power of the two uncorrelated models. 

\subsection{Sample size effect}
\begin{figure}[H]
\centering \label{fig:effect}
\includegraphics[width=0.8\textwidth]{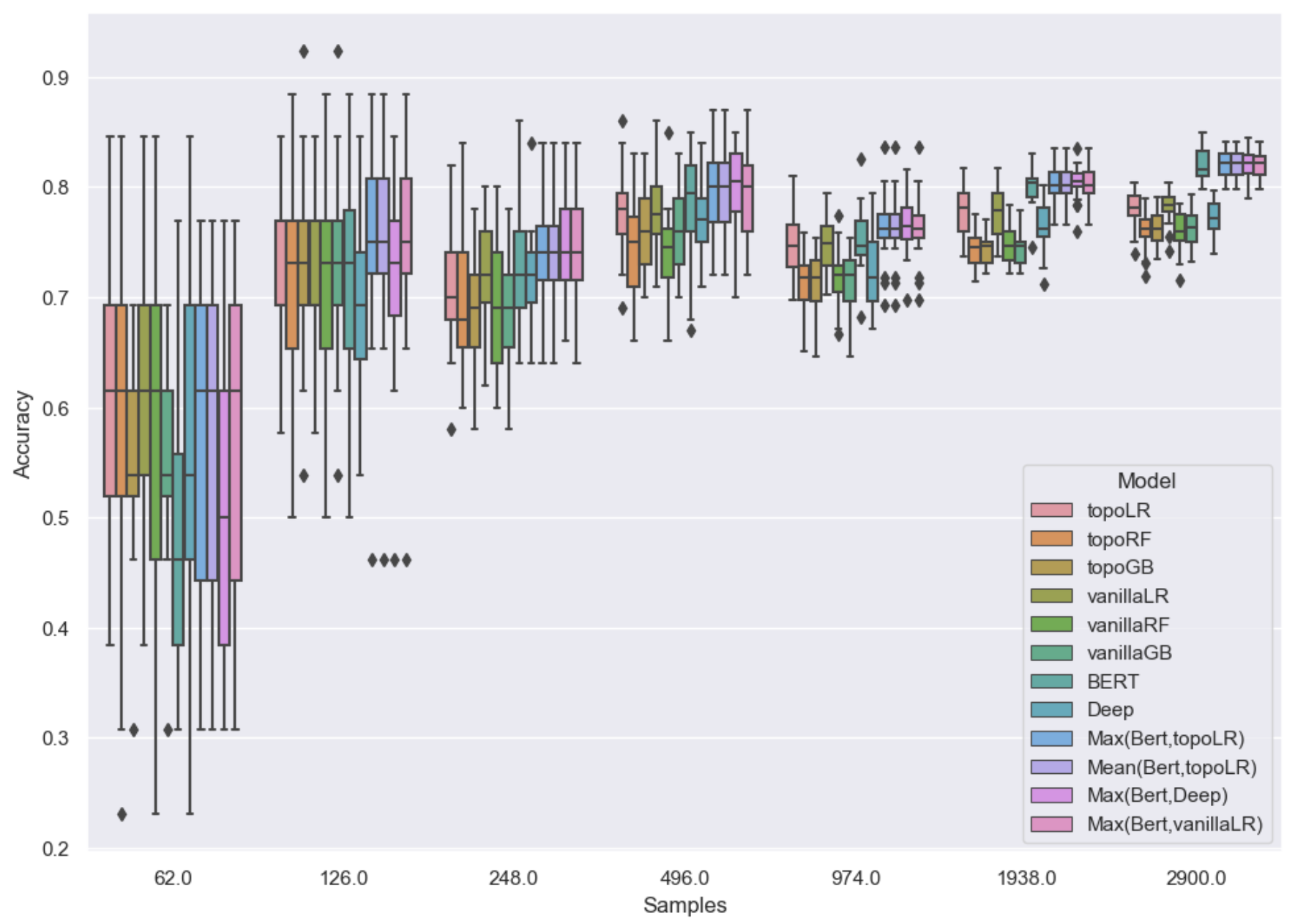}
\caption{\label{fig2}Box plots of the cross-validated performance of the different models at different data abundance regimes.}
\end{figure}

Let us now assess how the different models behave according to the sample size, i.e the number of tweets available for training. To get an accurate assessment of the performance, we cross-validated the results using 20 folds. In Figure \ref{fig2} we showcase the impact of sample size on the different methods. During the training and testing of all models, we used data from the \texttt{George Floyd Protest}.

This plot is showing that, in the low-data regime, large models like BERT are not as performing as the gradient-based methods.
On the other hand, in the high-data regime, BERT unleashes the power of his billions parameters and the results are stable and with high accuracy. 

Note that those results are validated in-sample (i.e. on the same event). In Section \ref{sec:out_sample}, we asses how well the models generalize out of the sample. 

\subsection{Runtime}
For many applications, runtime considerations play an important role. This section provides an overview of how long it takes each model to train on two different sample sizes. We conducted these measurements under the same conditions across all models. (The tests were conducted on a 3 CPU cores and 24 GB of RAM.) As shown in the following table, the times are expressed in seconds:
\begin{table}[H]
\centering 
\caption{\label{tab-runtime}Models Runtime}

\begin{tabular}{|c|c|c|c|c|c|c|c|c|} 

\hline
Sample Size  & TopoLR & TopolRF & TopolGB & BERT  & Deep & VanillaLR &  VanillaRF & VanillaGB \\
\hline
500 & 29.92 & 45.10 & 31.49 & 421.59 & 0.34 & 0.13 & 13.05 & 1.51\\
1k & 756.28 & 1044.44 & 924.99 & 2233.28 & 1.47 & 1.59 & 99.45 & 10.07 \\
\hline
\end{tabular}
\end{table}
According to the table, "VanillaLR" and "Deep" are the fastest models. Generally speaking, vanilla gradient methods are faster than topological methods. Finally, the slowest model is "BERT". 

\subsection{The test setups}
A good level of accuracy is not the only requirement of an Early Warning Signal detector. Indeed, we do not have any prior information or data regarding a completely new event when we wish to predict it. To account for all possible scenarios our model must be able to generalize in low data regime as well as in high data regime. Furthermore, it must be able to generalize well from one event to the other with little to no training.
Consequently, we have developed four different scenarios to test our proposed model as well as classical models. We refer to scenarios that use low and high sample sizes as \textit{low} and \textit{high data regimes}, respectively. When a model is tested on the same event it was trained on, we call it \textit{in-sample} evaluation, when tested on a different event we call it \textit{out-of-sample} evaluation. The datasets used in-sample evaluations are disjoint from the actual training data. 

\subsection{Low data regimes - in-sample}
In the low data regimes, the gradient methods seem to be the most effective ones. In the plot below we have performed  5-fold cross validation to assess the accuracy. The \texttt{George Floyd Protest} data was used to train and test all models.
The gradient methods have been found to be superior in almost all of the other events we have tested. See also \ref{subsection:Low-In}. 
A point worth mentioning is that we balanced our training set for all trials. This is due to the fact that imbalance affects the performance of classifiers.

\begin{figure}[H]
\centering
\includegraphics[width=0.6\textwidth]{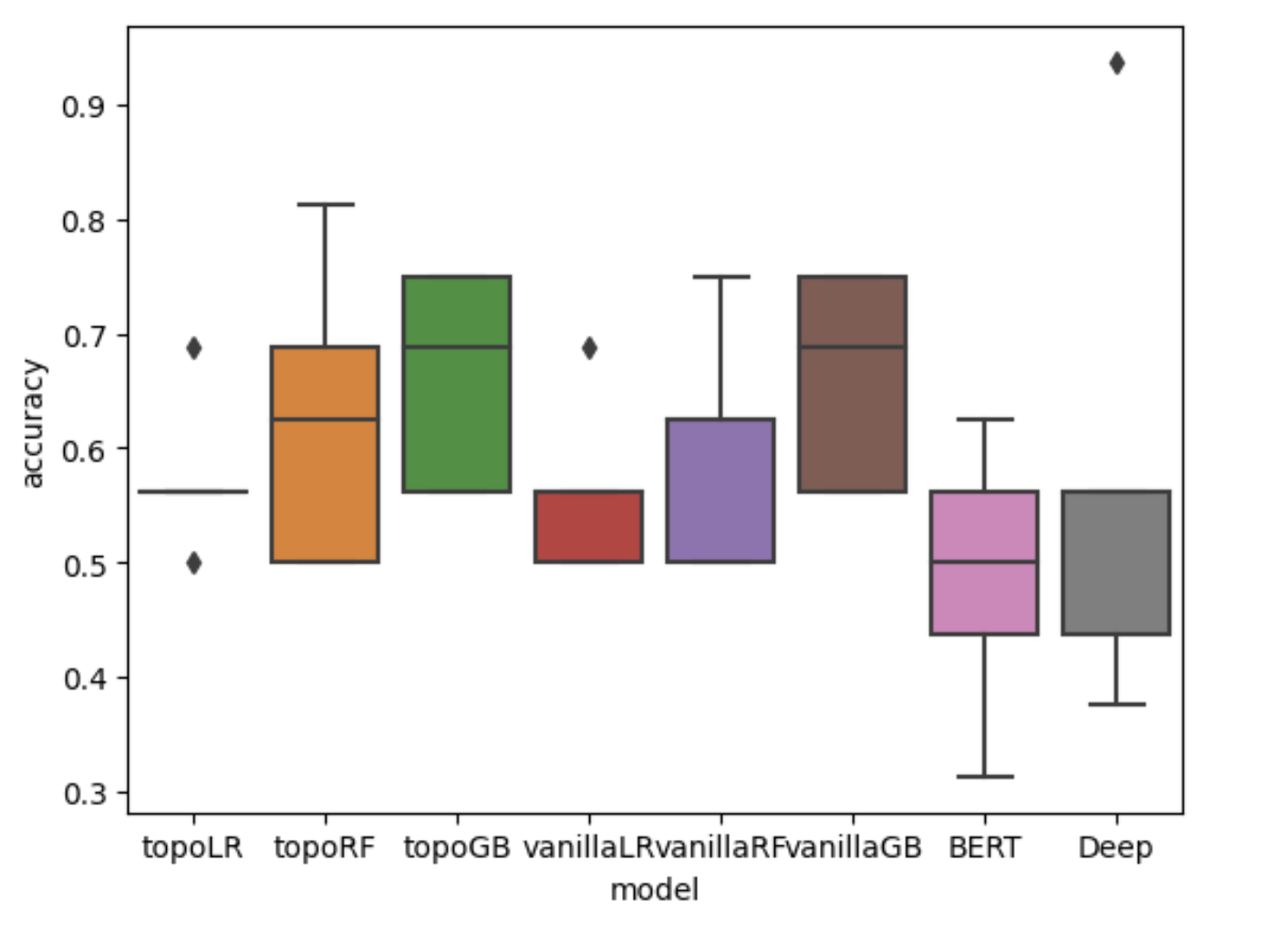}
\caption{\label{fig6}the cross-validated in-sample performance of the different models in the low-data regime for the \texttt{George Floyd Protest}.}
\end{figure}

\subsection{Low data regimes - out-of-sample}
We trained all models on a dataset pertaining to \textbf{black lives matter} (a.k.a. the \texttt{George Floyd Protest}) and then tested them on other events to understand the behavior of different models for the Low Regime Out sample. In this section, we test the trained models on the \texttt{Gun Control Protest} dataset. The results are as follows:

\begin{table}[H]
\parbox{.3\linewidth}{
\centering
\resizebox{.3\textwidth}{!}{%
\begin{tabular}{ccccc}
\hline
 & precision  &  recall & f1-score  & support \\
\hline
           0   &    0.63    &  0.73    &  0.68   &     26 \\
           1    &   0.77    &  0.68   &   0.72    &    34 \\
    accuracy   &           &          &      0.70   &   60\\
   macro avg   &    0.70   &  0.70    &  0.70    &    60 \\
weighted avg   &    0.71   &   0.70   &   0.70   &     60 \\
\hline
\end{tabular}
}
\caption{Topological RF}
}
\hfill
\parbox{.3\linewidth}{
\centering
\resizebox{.3\textwidth}{!}{%
\begin{tabular}{ccccc}
\hline
 & precision  &  recall & f1-score  & support \\
\hline
           0   &     0.70  &    0.66  &    0.68  &      32 \\
           1    &  0.63    &  0.68    &  0.66    &    28 \\
    accuracy   &           &          &      0.67   &     60 \\
   macro avg   &   0.67   &   0.67    &  0.67    &    60\\
weighted avg   &   0.67  &    0.67  &    0.67     &   60 \\
\hline
\end{tabular}
}
\caption{Topological GB}
}
\hfill
\parbox{.3\linewidth}{
\centering
\resizebox{.3\textwidth}{!}{%
\begin{tabular}{ccccc}
\hline
 & precision  &  recall & f1-score  & support \\
\hline
           0   &    0.70   &   0.72   &   0.71   &     29 \\
           1    &    0.73  &    0.71  &    0.72  &      31 \\
    accuracy   &           &          &  0.72   &     60 \\
   macro avg   &    0.72   &   0.72  &    0.72   &     60 \\
weighted avg   &     0.72  &    0.72   &   0.72    &    60 \\
\hline
\end{tabular}
}
\caption{Topological LR}
}
\end{table}

\begin{table}[H]
\parbox{.3\linewidth}{
\centering
\resizebox{.3\textwidth}{!}{%
\begin{tabular}{ccccc}
\hline
 & precision  &  recall & f1-score  & support \\
\hline
           0   &    0.63  &    0.76   &   0.69   &     25\\
           1    &  0.80   &   0.69    &  0.74    &    35 \\
    accuracy   &           &          &   0.72   &     60\\
   macro avg   &    0.72   &   0.72   &   0.71  &      60 \\
weighted avg   &   0.73    &  0.72   &   0.72   &     60 \\
\hline
\end{tabular}
}
\caption{Vanilla RF}
}
\hfill
\parbox{.3\linewidth}{
\centering
\resizebox{.3\textwidth}{!}{%
\begin{tabular}{ccccc}
\hline
 & precision  &  recall & f1-score  & support \\
\hline
           0   &    0.67  &    0.67    &  0.67     &   30 \\
           1    &   0.67   &   0.67     & 0.67      &  30 \\
    accuracy   &           &          &   0.67    &    60 \\
   macro avg   &    0.67   &   0.67   &   0.67    &    60 \\
weighted avg   &   0.67     &  0.67    &  0.67     &   60 \\
\hline
\end{tabular}
}
\caption{Vanilla GB}
}
\hfill
\parbox{.3\linewidth}{
\centering
\resizebox{.3\textwidth}{!}{%
\begin{tabular}{ccccc}
\hline
 & precision  &  recall & f1-score  & support \\
\hline
           0   &     0.70  &    0.72   &   0.71    &    29\\
           1    &   0.73   &   0.71   &   0.72     &   31 \\
    accuracy   &           &          &     0.72    &    60\\
   macro avg   &    0.72   &   0.72    &  0.72   &     60 \\
weighted avg   &    0.72  &    0.72     & 0.72    &    60 \\
\hline
\end{tabular}
}
\caption{Vanilla LR}
}
\hfill
\end{table}

\begin{table}[H]
\parbox{.45\linewidth}{
\centering
\resizebox{.45\textwidth}{!}{%
\begin{tabular}{ccccc}
\hline
 & precision  &  recall & f1-score  & support \\
\hline
           0   &      0.50    &    0.68  &      0.58   &       22 \\
           1    &    0.77     &   0.61     &   0.68      &    38 \\
    accuracy   &           &          &    0.63      &    60\\
   macro avg   &    0.63   &     0.64     &   0.63    &      60 \\
weighted avg   &     0.67   &     0.63   &     0.64    &      60 \\
\hline
\end{tabular}
}
\caption{Deep model}
}
\hfill
\parbox{.45\linewidth}{
\centering
\resizebox{.45\textwidth}{!}{%
\begin{tabular}{ccccc}
\hline
 & precision  &  recall & f1-score  & support \\
\hline
           0   &    0.63    &    0.68   &     0.66     &     28 \\
           1    &   0.70    &    0.66    &    0.68     &     32 \\
    accuracy   &           &          &   0.67    &      60 \\
   macro avg   &    0.67    &    0.67    &    0.67      &    60 \\
weighted avg   &    0.67    &    0.67    &    0.67     &     60 \\
\hline
\end{tabular}
}
\caption{BERT}
}
\end{table}
Accordingly to the results shown above, the gradient method works better than BERT in this scenario. The same conclusions can be drawn also when testing the models on other events: see \ref{subsection:Low-Out} for more details.

\subsection{High data regime - in-sample performance}

The following tables provide the in-sample performance (in validation) of the models in a high data regime. The protest analyzed is the \texttt{George Floyd Protest}. The dataset has been perfectly balanced: hence there are as many protest tweets as there are random tweets.

The results shown below have been obtained on a validation set, completely hidden in the training phase of the models.

\begin{table}[H]
\parbox{.3\linewidth}{
\centering
\resizebox{.3\textwidth}{!}{%
\begin{tabular}{ccccc}
\hline
 & precision  &  recall & f1-score  & support \\
\hline
           0   &    1.00  &    0.51    &  0.68   &   4709 \\
           1    &   0.04   &   1.00   &   0.08   &    101 \\
    accuracy   &           &          &   0.52   &   4810 \\
   macro avg   &    0.52   &   0.76   &   0.38   &   4810 \\
weighted avg   &    0.98   &   0.52   &   0.66   &   4810 \\
\hline
\end{tabular}
}
\caption{Topological RF}
}
\hfill
\parbox{.3\linewidth}{
\centering
\resizebox{.3\textwidth}{!}{%
\begin{tabular}{ccccc}
\hline
 & precision  &  recall & f1-score  & support \\
\hline
           0   &    0.96  &    0.59    &  0.73   &   3890 \\
           1    &   0.34   &   0.88   &   0.49   &    920 \\
    accuracy   &           &          &   0.65   &   4810 \\
   macro avg   &    0.65   &   0.74   &   0.61   &   4810 \\
weighted avg   &    0.84   &   0.65   &   0.68   &   4810 \\
\hline
\end{tabular}
}
\caption{Topological GB}
}
\hfill
\parbox{.3\linewidth}{
\centering
\resizebox{.3\textwidth}{!}{%
\begin{tabular}{ccccc}
\hline
 & precision  &  recall & f1-score  & support \\
\hline
           0   &    1.00  &    0.54    &  0.70   &   4447 \\
           1    &   0.15   &   1.00   &   0.26   &    363 \\
    accuracy   &           &          &   0.58   &   4810 \\
   macro avg   &    0.58   &   0.77   &   0.48   &   4810 \\
weighted avg   &    0.94   &   0.58   &   0.67   &   4810 \\
\hline
\end{tabular}
}
\caption{Topological LR}
}
\end{table}

\begin{table}[H]
\parbox{.3\linewidth}{
\centering
\resizebox{.3\textwidth}{!}{%
\begin{tabular}{ccccc}
\hline
 & precision  &  recall & f1-score  & support \\
\hline
           0   &    1.00  &    0.50    &  0.67   &   4810 \\
           1    &   0.0   &   0.00   &   0.0   &    0 \\
    accuracy   &           &          &   0.50   &   4810 \\
   macro avg   &    0.50   &   0.25   &   0.33   &   4810 \\
weighted avg   &    1.00   &   0.50   &   0.67   &   4810 \\
\hline
\end{tabular}
}
\caption{Vanilla RF}
}
\hfill
\parbox{.3\linewidth}{
\centering
\resizebox{.3\textwidth}{!}{%
\begin{tabular}{ccccc}
\hline
 & precision  &  recall & f1-score  & support \\
\hline
           0   &    1.00  &    0.50    &  0.67   &   4810 \\
           1    &   0.0   &   0.00   &   0.0   &    0 \\
    accuracy   &           &          &   0.50   &   4810 \\
   macro avg   &    0.50   &   0.25   &   0.33   &   4810 \\
weighted avg   &    1.00   &   0.50   &   0.67   &   4810 \\
\hline
\end{tabular}
}
\caption{Vanilla GB}
}
\hfill
\parbox{.3\linewidth}{
\centering
\resizebox{.3\textwidth}{!}{%
\begin{tabular}{ccccc}
\hline
 & precision  &  recall & f1-score  & support \\
\hline
           0   &    0.55  &    0.87    &  0.67   &   1522 \\
           1    &   0.92   &   0.67   &   0.77   &    3288 \\
    accuracy   &           &          &   0.73   &   4810 \\
   macro avg   &    0.73   &   0.77   &   0.72   &   4810 \\
weighted avg   &    0.80   &   0.73   &   0.74   &   4810 \\
\hline
\end{tabular}
}
\caption{Vanilla LR}
}
\hfill
\end{table}

\begin{table}[H]
\parbox{.45\linewidth}{
\centering
\resizebox{.45\textwidth}{!}{%
\begin{tabular}{ccccc}
\hline
 & precision  &  recall & f1-score  & support \\
\hline
           0   &    0.80  &    0.77    &  0.78   &   2503 \\
           1    &   0.76   &   0.79   &   0.78   &   2307 \\
    accuracy   &           &          &   0.78   &   4810 \\
   macro avg   &    0.78   &   0.78   &   0.78   &   4810 \\
weighted avg   &    0.78   &   0.78   &   0.78   &   4810 \\
\hline
\end{tabular}
}
\caption{Deep model}
}
\hfill
\parbox{.45\linewidth}{
\centering
\resizebox{.45\textwidth}{!}{%
\begin{tabular}{ccccc}
\hline
 & precision  &  recall & f1-score  & support \\
\hline
           0   &    0.85  &    0.82    &  0.84   &   2492 \\
           1    &   0.81   &   0.85   &   0.83   &   2318 \\
    accuracy   &           &          &   0.83   &   4810 \\
   macro avg   &    0.83   &   0.83   &   0.83   &   4810 \\
weighted avg   &    0.83   &   0.83   &   0.83   &   4810 \\
\hline
\end{tabular}
}
\caption{BERT}
}
\end{table}

The results shows that in-sample and in the high-data regime, BERT outperforms gradient based methods, this could be explained by the capacity of BERT of learning complex relations in the tweets when enough samples are provided. Additionally, we performed cross-validation for this scenario on a different event in \ref{subsection:High-In}.

In the next section, we assess how well the models generalize out of sample.

\subsection{High data regime - out-of-sample performance}\label{sec:out_sample}

Here below we showcase the out-of-sample performance of the different models considered (models trained on the \texttt{George Floyd Protest}): the protest analyzed is the \texttt{March for Our Lives}. Also in this case, the tweets have been perfectly balanced.

\begin{table}[H]
\parbox{.3\linewidth}{
\centering
\resizebox{.3\textwidth}{!}{%
\begin{tabular}{ccccc}
\hline
 & precision  &  recall & f1-score  & support \\
\hline
           0   &    1.00  &    0.50    &  0.67   &   4299 \\
           1    &   0.00   &   0.00   &   0.00   &   1 \\
    accuracy   &           &          &   0.50   &   4300 \\
   macro avg   &    0.50   &   0.25   &   0.33   &   4300 \\
weighted avg   &    1.00   &   0.50   &   0.67   &   4300 \\
\hline
\end{tabular}
}
\caption{Topological RF}
}
\hfill
\parbox{.3\linewidth}{
\centering
\resizebox{.3\textwidth}{!}{%
\begin{tabular}{ccccc}
\hline
 & precision  &  recall & f1-score  & support \\
\hline 
           0   &    0.95  &    0.60    &  0.74   &   3389 \\
           1    &   0.37   &   0.89   &   0.53   &   902 \\
    accuracy   &           &          &   0.66   &   4300 \\
   macro avg   &    0.66   &   0.75   &   0.63   &   4300 \\
weighted avg   &    0.83   &   0.66   &   0.69   &   4300 \\
\hline
\end{tabular}
}
\caption{Topological GB}
}
\hfill
\parbox{.3\linewidth}{
\centering
\resizebox{.3\textwidth}{!}{%
\begin{tabular}{ccccc}
\hline
 & precision  &  recall & f1-score  & support \\
\hline
           0   &    1.00  &    0.51    &  0.67   &   4238 \\
           1    &   0.03   &   0.95   &   0.05   &   62 \\
    accuracy   &           &          &   0.51   &   4300 \\
   macro avg   &    0.51   &   0.73   &   0.36   &   4300 \\
weighted avg   &    0.98   &   0.51   &   0.66   &   4300 \\
\hline
\end{tabular}
}
\caption{Topological LR}
}
\end{table}

\begin{table}[H]
\parbox{.3\linewidth}{
\centering
\resizebox{.3\textwidth}{!}{%
\begin{tabular}{ccccc}
\hline
 & precision  &  recall & f1-score  & support \\
\hline
           0   &    1.00  &    0.50    &  0.67   &   4299 \\
           1    &   0.00   &   0.00   &   0.00   &   1 \\
    accuracy   &           &          &   0.50   &   4300 \\
   macro avg   &    0.50   &   0.25   &   0.33   &   4300 \\
weighted avg   &    1.00   &   0.50   &   0.67   &   4300 \\
\hline
\end{tabular}
}
\caption{Vanilla RF}
}
\hfill
\parbox{.3\linewidth}{
\centering
\resizebox{.3\textwidth}{!}{%
\begin{tabular}{ccccc}
\hline
 & precision  &  recall & f1-score  & support \\
\hline
           0   &    1.00  &    0.50    &  0.67   &   4299 \\
           1    &   0.00   &   0.00   &   0.00   &   1 \\
    accuracy   &           &          &   0.50   &   4300 \\
   macro avg   &    0.50   &   0.25   &   0.33   &   4300 \\
weighted avg   &    1.00   &   0.50   &   0.67   &   4300 \\
\hline
\end{tabular}
}
\caption{Vanilla GB}
}
\hfill
\parbox{.3\linewidth}{
\centering
\resizebox{.3\textwidth}{!}{%
\begin{tabular}{ccccc}
\hline
 & precision  &  recall & f1-score  & support \\
\hline
           0   &    0.55  &    0.86    &  0.67   &   1370 \\
           1    &   0.91   &   0.67   &   0.77   &   2930 \\
    accuracy   &           &          &   0.73   &   4300 \\
   macro avg   &    0.73   &   0.77   &   0.72   &   4300 \\
weighted avg   &    0.80   &   0.73   &   0.74   &   4300 \\
\hline
\end{tabular}
}
\caption{Vanilla LR}
}
\end{table}


\begin{table}[H]
\parbox{.45\linewidth}{
\centering
\resizebox{.45\textwidth}{!}{%
\begin{tabular}{ccccc}
\hline
 & precision  &  recall & f1-score  & support \\
\hline
           0    &   0.80   &   0.73   &   0.77   &   2359 \\
           1   &    0.71  &    0.78    &  0.74   &   1941 \\
    accuracy   &           &          &   0.75   &   4300 \\
   macro avg   &    0.75   &   0.76   &   0.75   &   4300 \\
weighted avg   &    0.76   &   0.75   &   0.75   &   4300 \\
\hline
\end{tabular}
}
\caption{Deep model}
}
\hfill
\parbox{.45\linewidth}{
\centering
\resizebox{.45\textwidth}{!}{%
\begin{tabular}{ccccc}
\hline
 & precision  &  recall & f1-score  & support \\
\hline
           0    &   0.84   &   0.79   &   0.82   &   2282 \\
           1   &    0.78  &    0.83    &  0.80   &   2018 \\
    accuracy   &           &          &   0.81   &   4300 \\
   macro avg   &    0.81   &   0.81   &   0.81   &   4300 \\
weighted avg   &    0.81   &   0.81   &   0.81   &   4300 \\
\hline
\end{tabular}
}
\caption{BERT}
}
\end{table}
In the above, we see that BERT generalizes the best. It should be noted, however, that this is not true for other events: see \ref{subsection:High-Out} for more details. 
Let us then assess how it performs on a slightly different type of event (not of class "protest", as above). We will hence consider a different event for the next out-of-sample analysis: the \texttt{Ukrainian war}.

\begin{table}[H]
\parbox{.3\linewidth}{
\centering
\resizebox{.3\textwidth}{!}{%
\begin{tabular}{ccccc}
\hline
 & precision  &  recall & f1-score  & support \\
\hline
           0    &   0.86   &   0.69   &   0.77   &   3088 \\
           1   &    0.62  &    0.82    &  0.70   &   1862 \\
    accuracy   &           &          &   0.74   &   4950 \\
   macro avg   &    0.74   &   0.76   &   0.74   &   4950 \\
weighted avg   &    0.77   &   0.74   &   0.74   &   4950 \\
\hline
\end{tabular}
}
\caption{Topological RF}
}
\hfill
\parbox{.3\linewidth}{
\centering
\resizebox{.3\textwidth}{!}{%
\begin{tabular}{ccccc}
\hline
 & precision  &  recall & f1-score  & support \\
\hline
           0    &   0.83   &   0.61   &   0.70   &   3363 \\
           1   &    0.47  &    0.73    &  0.57   &   1587 \\
    accuracy   &           &          &   0.65   &   4950 \\
   macro avg   &    0.65   &   0.67   &   0.63   &   4950 \\
weighted avg   &    0.71   &   0.65   &   0.66   &   4950 \\
\hline
\end{tabular}
}
\caption{Topological GB}
}
\hfill
\parbox{.3\linewidth}{
\centering
\resizebox{.3\textwidth}{!}{%
\begin{tabular}{ccccc}
\hline
 & precision  &  recall & f1-score  & support \\
\hline
           0    &   0.83   &   0.55   &   0.66   &   3771 \\
           1   &    0.31  &    0.65    &  0.42   &   1179 \\
    accuracy   &           &          &   0.57   &   4950 \\
   macro avg   &    0.57   &   0.60   &   0.54   &   4950 \\
weighted avg   &    0.71   &   0.57   &   0.60   &   4950 \\
\hline
\end{tabular}
}
\caption{Topological LR}
}
\end{table}

\begin{table}[H]
\parbox{.3\linewidth}{
\centering
\resizebox{.3\textwidth}{!}{%
\begin{tabular}{ccccc}
\hline
 & precision  &  recall & f1-score  & support \\
\hline
           0    &   0.87   &   0.69   &   0.77   &   3094 \\
           1   &    0.62  &    0.82    &  0.71   &   1856 \\
    accuracy   &           &          &   0.74   &   4950 \\
   macro avg   &    0.74   &   0.76   &   0.74   &   4950 \\
weighted avg   &    0.77   &   0.74   &   0.75   &   4950 \\
\hline
\end{tabular}
}
\caption{Vanilla RF}
}
\hfill
\parbox{.3\linewidth}{
\centering
\resizebox{.3\textwidth}{!}{%
\begin{tabular}{ccccc}
\hline
 & precision  &  recall & f1-score  & support \\
\hline
           0    &   0.82   &   0.60   &   0.70   &   3379 \\
           1   &    0.46  &    0.72    &  0.56   &   1571 \\
    accuracy   &           &          &   0.64   &   4950 \\
   macro avg   &    0.64   &   0.66   &   0.63   &   4950 \\
weighted avg   &    0.71   &   0.64   &   0.65   &   4950 \\
\hline
\end{tabular}
}
\caption{Vanilla GB}
}
\hfill
\parbox{.3\linewidth}{
\centering
\resizebox{.3\textwidth}{!}{%
\begin{tabular}{ccccc}
\hline
 & precision  &  recall & f1-score  & support \\
\hline
           0    &   0.83   &   0.54   &   0.66   &   3780 \\
           1   &    0.31  &    0.65    &  0.41   &   1170 \\
    accuracy   &           &          &   0.57   &   4950 \\
   macro avg   &    0.57   &   0.60   &   0.54   &   4950 \\
weighted avg   &    0.71   &   0.57   &   0.60   &   4950 \\
\hline
\end{tabular}
}
\caption{Vanilla LR}
}
\end{table}

\begin{table}[H]
\parbox{.45\linewidth}{
\centering
\resizebox{.45\textwidth}{!}{%
\begin{tabular}{ccccc}
\hline
 & precision  &  recall & f1-score  & support \\
\hline
           0    &   0.80   &   0.51   &   0.62   &   3896 \\
           1   &    0.23  &    0.53    &  0.32   &   1054 \\
    accuracy   &           &          &   0.51   &   4950 \\
   macro avg   &    0.51   &   0.52   &   0.47   &   4950 \\
weighted avg   &    0.68   &   0.51   &   0.56   &   4950 \\
\hline
\end{tabular}
}
\caption{Deep model}
}
\hfill
\parbox{.45\linewidth}{
\centering
\resizebox{.45\textwidth}{!}{%
\begin{tabular}{ccccc}
\hline
 & precision  &  recall & f1-score  & support \\
\hline
           0    &   0.85   &   0.49   &   0.62   &   4335 \\
           1   &    0.10  &    0.40    &  0.16   &   615 \\
    accuracy   &           &          &   0.48   &   4950 \\
   macro avg   &    0.48   &   0.44   &   0.39   &   4950 \\
weighted avg   &    0.76   &   0.48   &   0.56   &   4950 \\
\hline
\end{tabular}
}
\caption{BERT}
}
\end{table}

We see, in this example, that BERT generalizes less well than the gradient methods. Conceptually, the BERT and the gradient based methods have an important difference: Once trained, the BERT model considers each tweet individually without relation to other tweets. On the other hand, the gradient based methods by definition craft their features in relation to other tweets. We see in particular that BERT underestimates the number of event related tweets. The Deep feed forward model also bases its inference on the contents of an individual tweet and it exhibits similar underestimates, albeit to a lesser degree.

Our hypothesis is that perhaps BERT learns tweets of type 'domestic political protest'. The sentiment in the English language tweets on the Ukrainian war is possibly sufficiently different from a domestic event, where many of the tweeters may actually be participating in the events.

The gradient methods on the other hand are better suited for detecting anomalies of unknown kinds. This is because their inference is not particularly based on the specific content of an individual tweet, but rather on how the tweet relates to other tweets, that is, how the tweet is related to what else is happening in the world.
\subsection{Summary}
\begin{figure}[H]
\centering
\includegraphics[width=0.36\textwidth]{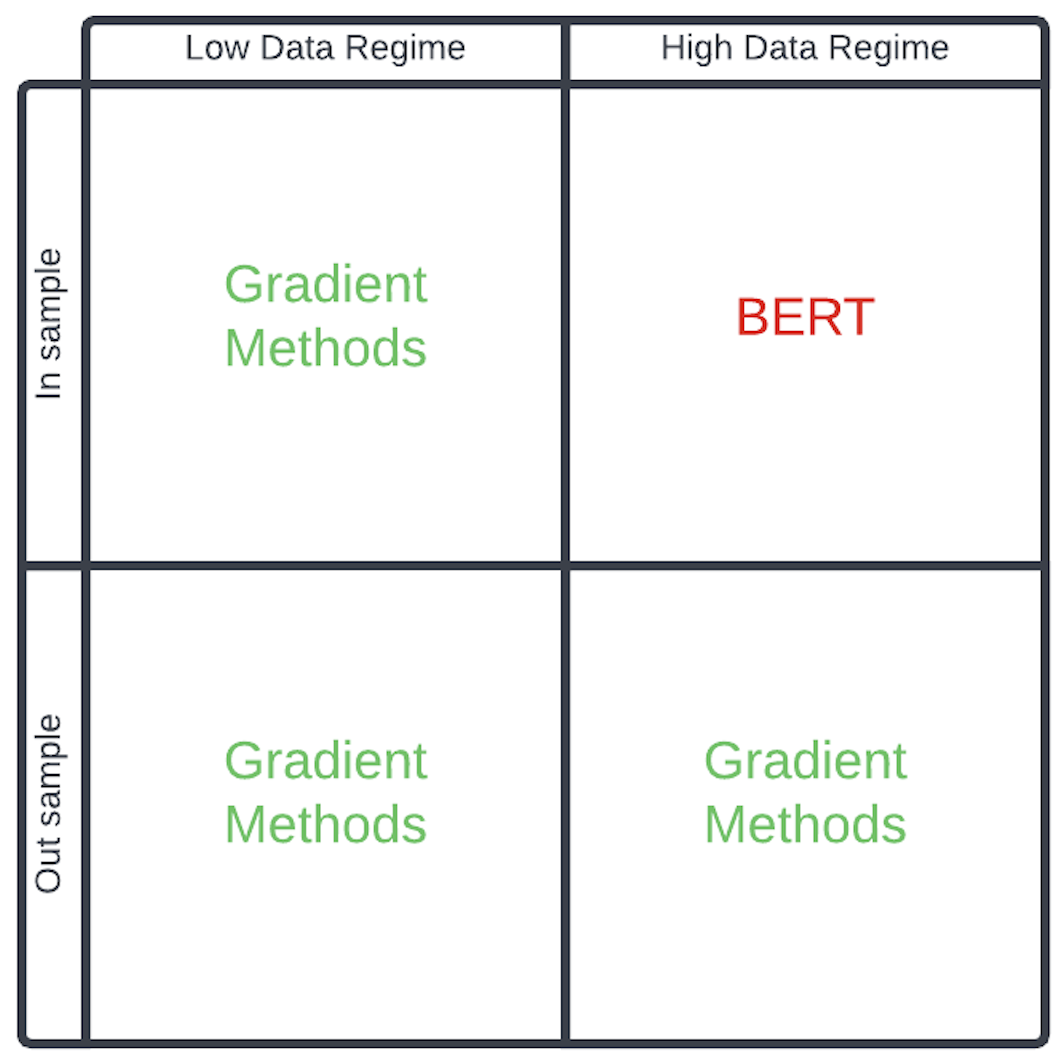}
\caption{\label{fig-res-summ} The best model class in each scenario}
\end{figure}

In this section, we provide a short summary of the above. Except for the high-data in-sample scenario, our gradient solutions perform better than state-of-the-art deep learning models (see Figure \ref{fig-res-summ}).

A good out-of-sample performance is crucial for a sensible monitoring tool of early warning signals, given the results of the previous sections we will now assess how gradient based methods perform on a time series scenario describing the evolution of the media resonance of an event. 

\section{Time series analysis}
In a real-world scenario, events go from a period of low media resonance to high resonance: the evolution can hence be described as a time series. 

For a fixed time window of one week, we compute the ratio of tweets that are protest-related versus the total number of retrieved tweets. This ratio forms a single point in our time series, whose time coordinate would be the last day of the considered interval; for each time window we run the topological algorithm (topological gradient-based method with a random forest at the top to detect the different classes) to compute the same ratio for the predicted labels. This is what is shown in the plots.
Furthermore, from the point of view of the model, this implies periods of highly imbalanced data.

\subsection{Artificial data}

Let us first assess how gradient-based model performs on an artificial time series modeling different percentages of event-related tweets. The models are trained on the \texttt{George Floyd protest} while the results are evaluated in the \texttt{Gun Control protest}. We designed different synthetic scenarios to test our model:

\begin{figure}[H]
    \centering
    \includegraphics[width=0.9\textwidth]{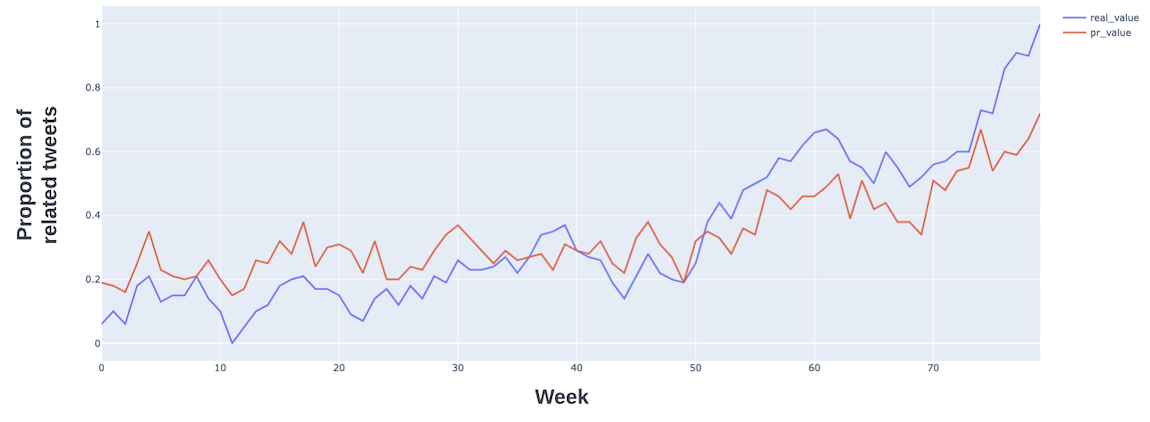} 
        \caption{Performance of our models on the \texttt{Gun Control protest} synthetic time series, 1st case. Predicted evolution in \textbf{red} vs true evolution in \textbf{blue}.}%
\end{figure}
\begin{figure}[H]
    \centering
    \includegraphics[width=0.9\textwidth]{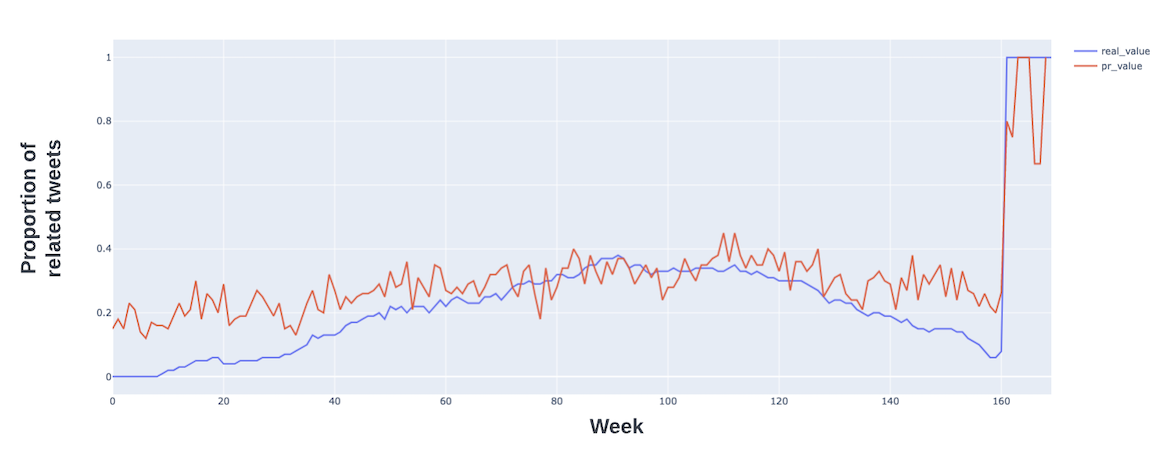}
    \caption{Performance of our models on the \texttt{Gun Control protest} synthetic time series, 2nd case.}%
\end{figure}
\begin{figure}[H]
    \centering
    \includegraphics[width=0.9\textwidth]{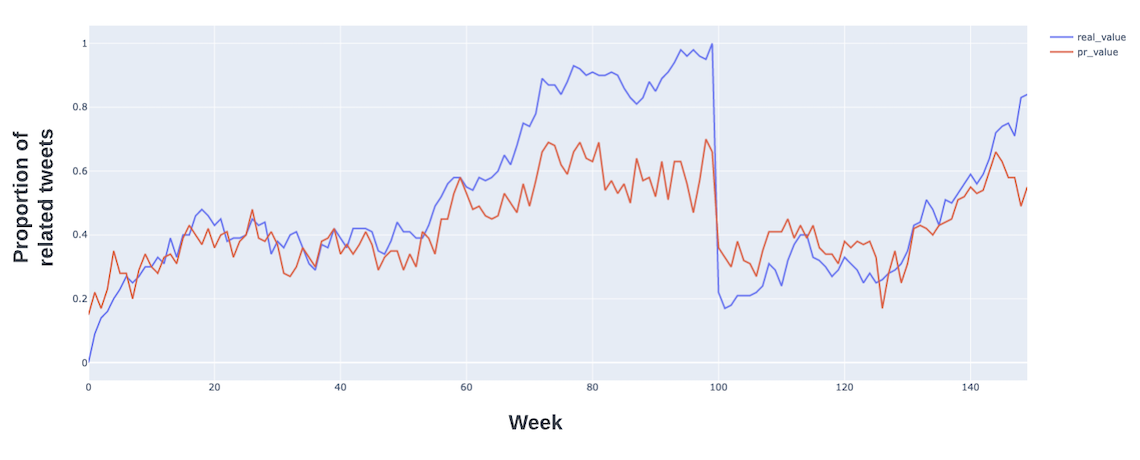}
    \caption{Performance of our models on the \texttt{Gun Control protest} synthetic time series, 3rd case.}%
\end{figure}
\begin{figure}[H]
    \centering
    \includegraphics[width=0.9\textwidth]{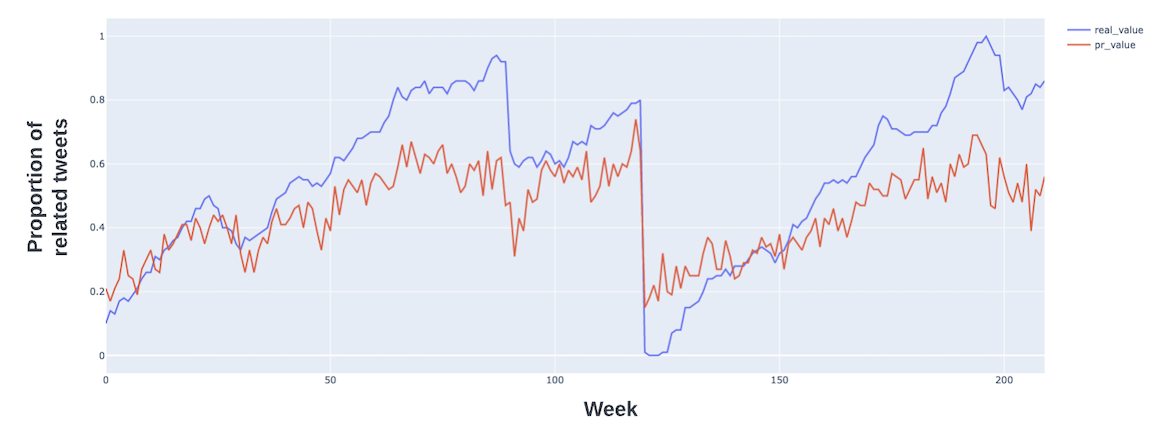} 
    \caption{Performance of our models on the \texttt{Gun Control protest} synthetic time series, 4th case.}%
\end{figure}
As can be seen in the figures, the topological method does generalize well to unseen events without being heavily affected by the dataset imbalance. Put differently, the predicted evolution of tweets is well correlated with the true evolution. Refer to \ref{subsection:artif-time-series} for more results.

\subsection{Real time series}
We will now extract real tweets over a defined period of time and assess how well topological methods work on a real-time series. 

\subsubsection{Fetching the data}
We used twitter API to fetch the data. However, due to the API limitation on the number of tweets, this is difficult to accomplish. To circumvent the limitation, we divide long time intervals into shorter intervals (of 6 hours). For each interval, we set a maximum number of 200 tweets and use the API to fetch tweets related to the event. We assume that we fetched all the event-related tweets if the number of tweets in the results of that interval is less than 200. We complete all the results of each interval with random tweets in order to reach 200 tweets. So, 200 tweets per 6 hours intervals constitute our final set of tweets. Finally, we aggregate the data on a weekly basis due to the noisy behavior that occurs in shorter intervals. 
\subsubsection{Results} 
In this section we show the results of our analysis, in which we train our gradient-based method on the \texttt{George Floyd Protest} and then test it out of sample on various types of events. 
\begin{itemize}
    \item 
    The result on the \texttt{Ukrainian war}:
        \begin{figure}[H]
        \centering
        \includegraphics[width=1.0\textwidth]{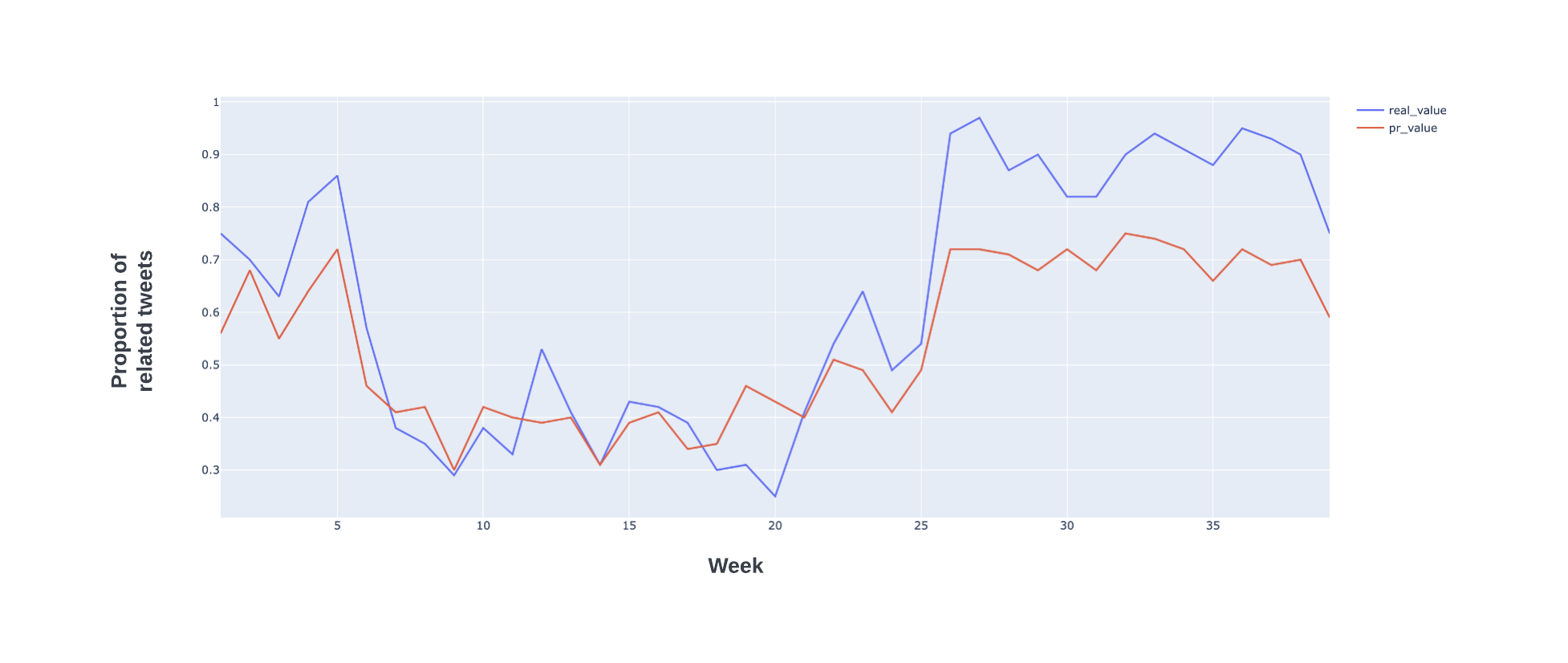}
        \caption{\label{real-timeseries-ukrine-war}The result of topological gradient method on time series related to the \texttt{Ukrainian war}. Predicted evolution in \textbf{red} vs true evolution in \textbf{blue}.}
        \end{figure}
    The predicted evolution shows a high correlation with real evolution. In particular, the model seems to have been able to capture the peak around week 25.
    \item
        The result on the \texttt{2018 Women's march} protest:
        \begin{figure}[H]
        \centering
        \includegraphics[width=1.0\textwidth]{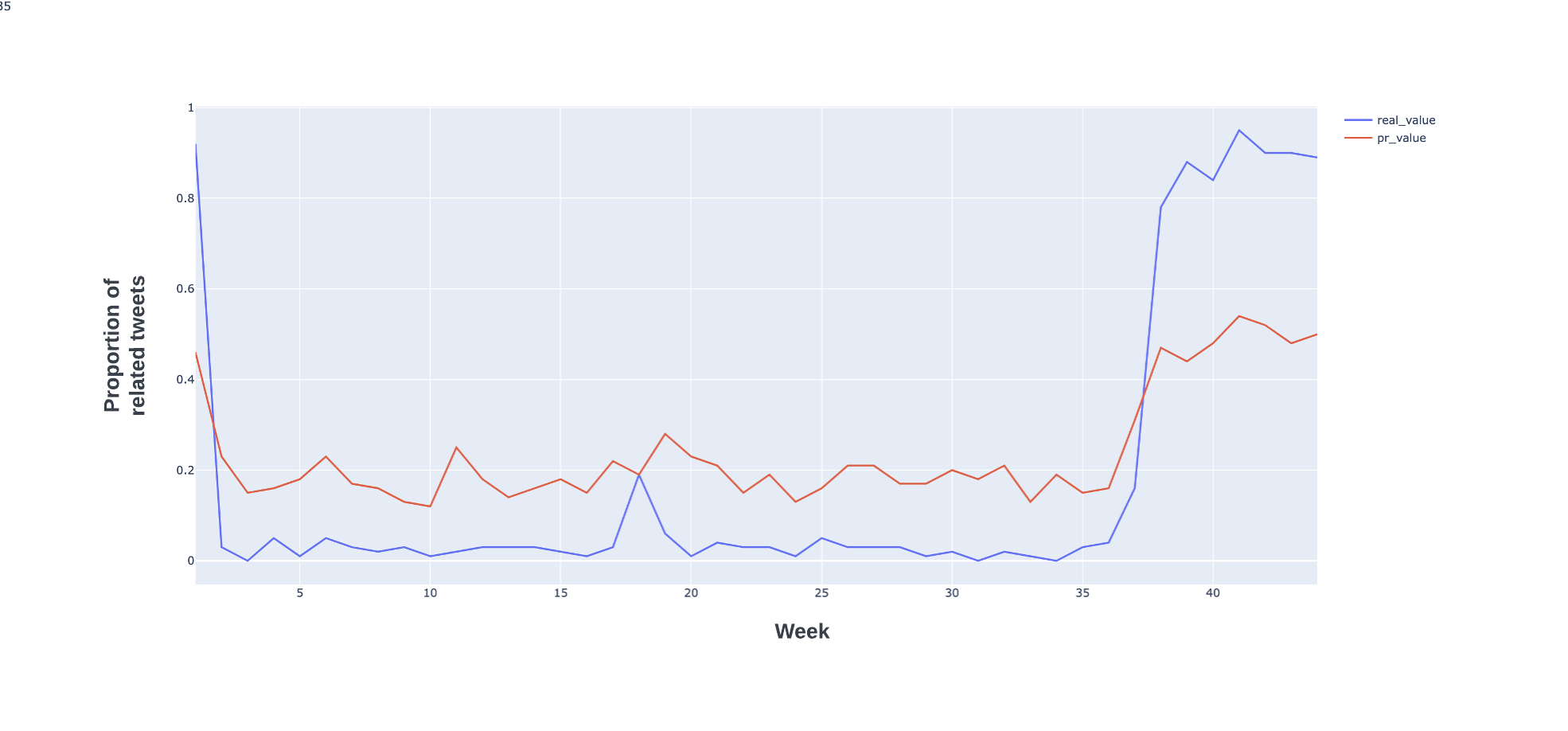}
        \caption{\label{real-timeseries-fem}The result of topological gradient method on time series related to \texttt{2018 Women's march} (The protest occurs approximately at \textbf{week 42}) }
        \end{figure}
   \item
        The result on the \texttt{March for Our Lives} protest:
        \begin{figure}[H]
        \centering
        \includegraphics[width=1.0\textwidth]{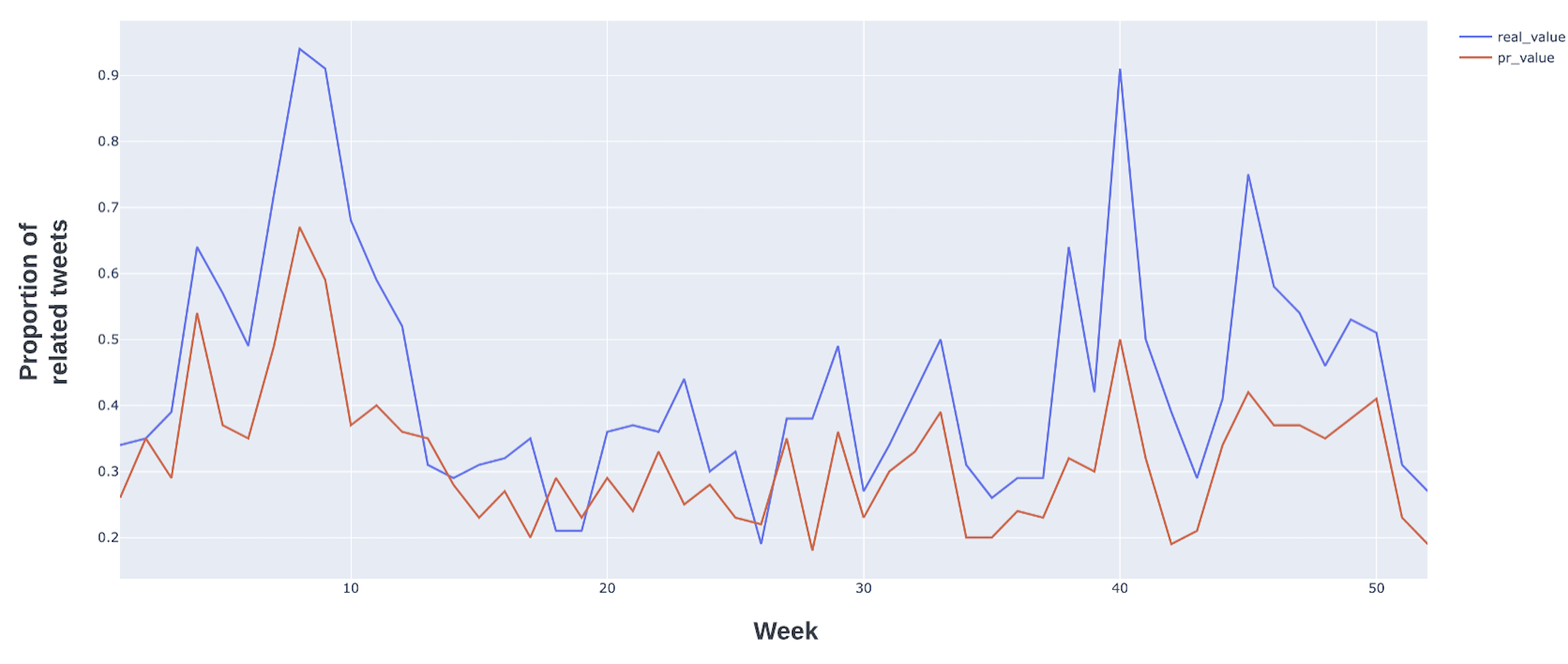}
        \caption{\label{real-timeseries-gun-cont}The result of topological gradient method on time series related \texttt{March for Our Lives}  protest (The protest occurs after \textbf{week 40}) }
        \end{figure}
    
\end{itemize}
It seems indeed that the predictions of our method highly correlate with the real-time series. Furthermore, our methods are capable of detecting the major peaks in the data.

\section{Conclusion and Future work}
The results obtained in this project show that gradient base methods seem to generalize well in a real time series scenario. In particular, it seems to be able to work out-of-sample on different events and in imbalanced conditions. Those results suggest that it might be possible to use our algorithms in an online monitoring tool that scans the streaming of tweets in real time. We should further empirically validate those results in more events covering a larger spectrum of scenarios. Furthermore, we could use the methods developed above to build an early warning indicator of protests rise and measure the sensitivity and specificity of such indicator.

\section*{Acknowledgement}
This research has been founded by armasuisse, contract number CYD-C-2021017, to whom the authors are grateful.

\renewcommand{\thesection}{A}
\section{Appendix}
In this appendix, we provide more details about the less well known algorithms used in this work.

\subsection{Sentence embedding}
\label{sec:pg2}
Some of the models and algorithms that we used in our analysis require the data to be provided as points. In order to transform entire tweets into points, we use pre-trained \texttt{sent2vec} models \citealp{moghadasi2020sent2vec}.
Sentence-to-vector or sentence-transformers models are a family of NLP (Natural Language Processing) models that map sentences to vectors. They get sentences as input and output a vector in Euclidean space with a given dimension. In this framework, we use "Bert-base-nli-mean-tokens": a pre-trained sentence-transformers model that maps sentences and paragraphs to a 768-dimensional dense vector space and can be used for tasks like clustering or classification. It is a modification of the pre-trained BERT network.\texttt{sent2vec} uses some unique network structure of BERT to derive semantically meaningful sentence embeddings that can be compared using cosine similarity. We use a python framework for state-of-the-art sentences called SBERT to implement it: \url{https://www.sbert.net/}

\subsection{Persistence gradient}
\label{sec:pg1}
The central object in topological data analysis is the persistence diagram \citealp{chazal2021introduction}. Such a diagram encodes the shape of a point cloud. Here below an example of a persistence diagram:

\begin{figure}[H]
\centering
\includegraphics[width=0.8\textwidth]{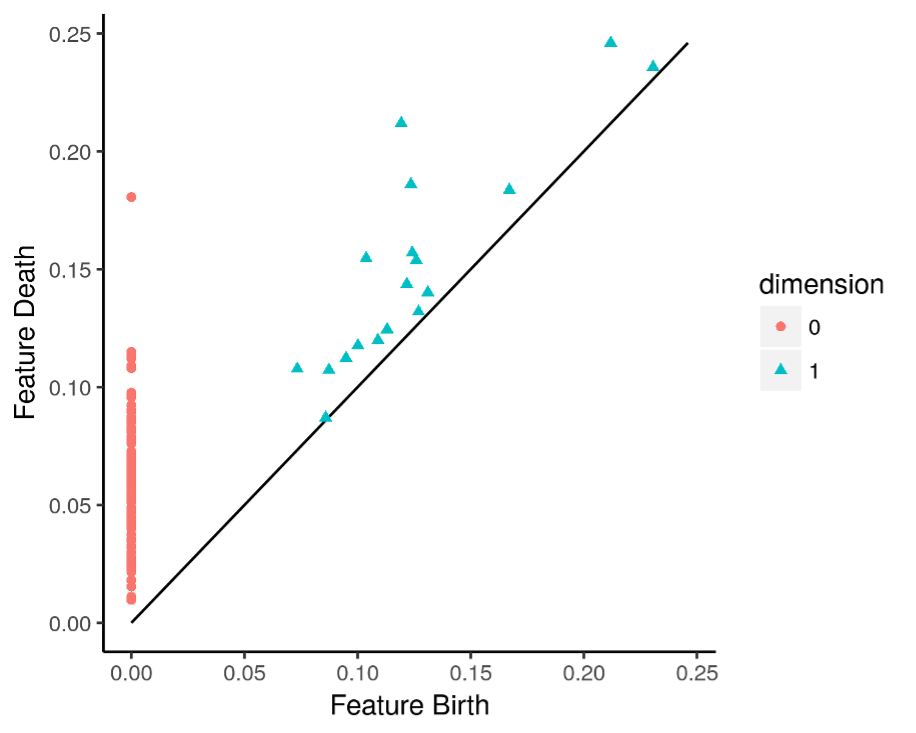}
\caption{\label{fig5}Example of a persistence diagram.}
\end{figure}

Each red point in the diagram above corresponds to a connected component of the point cloud: points are part of the same connected component if the distance between them is less than the \texttt{Feature Death} value. The blue triangles, on the other hand, represent loops within the point clouds\footnote{An equivalence class of loops, where the equivalence is given by homotopy equivalence.}. 
The persistence gradient is a technique to differentiate a loss function built out of the features present in the persistence diagram (i.e., birth and death features for all the points in the diagram) with respect to the position of the points of the point cloud \citealp{corcoran2020regularization}. 

Let us now set some notation:
\begin{enumerate}
    \item $K$ is the simplicial complex
    \item $p$ is the number of finite bars (generated by positive simplices and eliminated by negative simplices)
    \item $q$ is the number of infinitely persistent features.
\end{enumerate}

Clearly $|K| = 2p + q$.

We will focus on Vietoris-Rips (VR) persistence, as the other filtrations are done similarly.

The first step is to notice that a filtration can be seen as a map 
$$\Phi : A \to \mathbb{R}^{|K|},$$

where the space $A$ is the space of point clouds of $n$ elements of $d$ features each, hence isomorphic to $\mathbb{R}^{n \times d}$.

The map $\Phi$ is given explicitly: given $X$ a point cloud (i.e. an element of $A$),

$$ \Phi_\sigma(X) = max_{i,j \in \sigma}||x_i-x_j||$$

The above is the $\sigma$-th component of the whole $\Phi$. 
Here $\sigma$ is a simplex, i.e. a coordinate of $\mathbb{R}^{|K|}$.

A persistent diagram imposes:
\begin{itemize}
    \item The identification of $\mathbb{R}^{|K|}=(\mathbb{R}^{2})^{p} \times \mathbb{R}^{q}$
    \item The pairing of positive with negative simplices and the identification of unpaired positive simples.
\end{itemize}

Hence

$$Pers:Filt_K \subset \mathbb R^{|K|} \to (\mathbb R^2)^p \times \mathbb R^q, \Phi(X) \mapsto D = \cup_i^p (\Phi_{\sigma_{i_1}}(X) , \Phi_{\sigma_{i_2}}(X) ) \times \cup_j^q (\Phi_{\sigma_j}(X),+\infty).$$

where $|K|=2p + q$.

Finally, we can also define persistence functions: they are functions like $E: (\mathbb R^2)^p \times \mathbb R^q \to \mathbb R$, invariant under permutations of $p$ and $q$. Regarding this article's objective, persistence entropy (dimension zero) is defined as the topological gradient base method's loss function. Intuitively, this measures the entropy of the lifetime (death-birth) of points in persistent diagrams. Accordingly, D's persistence entropy is defined as follows:
$$E(D)=-\sum_{i}^p p_{i} \log \left(p_{i}\right)$$
where 
$$
p_{i}=\frac{\left(d_{i}-b_{i}\right)}{L_{D}} \quad \text{ and } \quad L_{D}=\sum_{i}^p \left(d_{i}-b_{i}\right)
$$
We can now define a loss function $L:= E.Pers.\Phi : A \to \mathbb R, A = (\mathbb R^d)^n$ as before. Can we compute the gradient of $L$ with respect to the point cloud? Observe that $Pers$ is merely a permutation of the coordinates, thus its partial derivatives w.r.t. the filtration are either 1 or 0.
Thus, since all the components of $L$ are differentiable, so is $L$ by Leibnitz rule.
One can thus implement it is \texttt{Pytorch} using \texttt{autograd}: we provide an implementation in \texttt{giotto-deep} (\cite{caorsi2022giotto}).
\paragraph{}
The algorithm with this loss function attempts to change the location of points so that they only have a single component, thus resulting in the lowest possible entropy. Consequently, the evolutionary vectors at the end of the algorithm may contain information about the class of the sample. 

\subsection{Impact of higher homology groups}

Persistence gradients have the possibility to decide what are the homology groups to be taken into account in the computation and consequently the nature of the geometric features. 

In this short section, we have analyzed the impact of the incorporating the homology groups of dimensions 1 and 2 into our analysis.

\begin{figure}[H]
\centering
\includegraphics[width=0.96\textwidth]{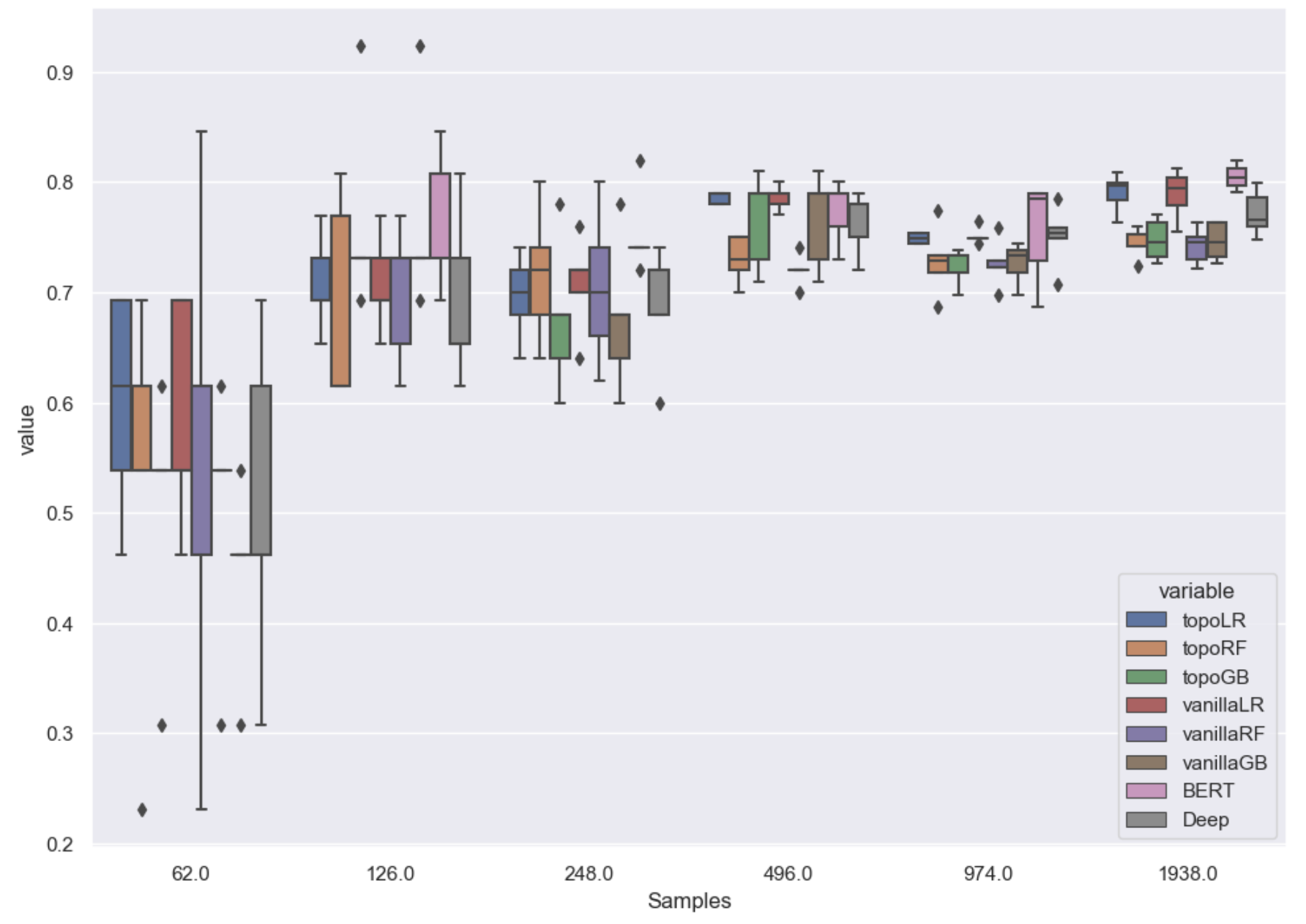}
\caption{\label{fig3}Box plots with the gradient methods for higher homology groups, $H_0$, $H_1$ and $H_2$.}
\end{figure}

The results are displayed in Figure \ref{fig3}.

Since the vanilla gradients (that are not affected by the enhancement of the topological features) and the persistent gradients always seem to give compatible results, we are driven to the conclusion that the higher topological features are not improving the classification power of our persistence-gradient-based method.

\subsection{Why Vanilla gradients correlates with topological gradients}
\label{subsection:Vanilla-Pers}
In Section \ref{sec:corr}, we have seen that the persistence gradient with RF and the vanilla gradient with RF are very correlated. Here is a likely explanation:
For this special task, persistent entropy is defined as the loss function of persistent gradient methods in dimension zero. The dimension zero of homology takes care of the number of connected components. Therefore, every generator of persistent homology in dimension zero represents a connected component that was born in the birth time and died (in other words got merged with some other connected components) at the death time. If we define the lifetime of a generator as the difference between its birth and death time, then we can calculate the entropy of the lifetimes of all generators. So this loss function tries to change the location of points in a way that they just have a single component which leads the entropy to have the minimum value, i.e one. Due to this logic, the vanilla gradient method accomplishes the same goal: by minimizing the distances between points with barycenter, the vanilla gradient also changes the location of points such that at the end of the algorithm, we only have one connected component. 

\bibliographystyle{johd}
\bibliography{bib}

\renewcommand{\thesection}{S}
\section{Supplementary Material}
\label{section:SM}

\subsection{Low data regimes - in-sample}
\label{subsection:Low-In}
In this section we show the results of the models trained and validated in the same events.
\begin{itemize}
    \item 
    \texttt{2017 Women's march} protest:
    \begin{figure}[H]
    \centering
    \includegraphics[width=0.6\textwidth]{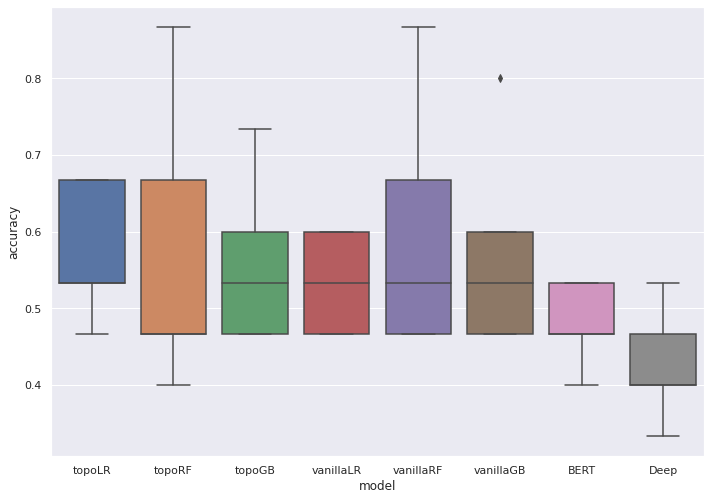}
    \end{figure}
    
    \item 
    \texttt{2018 Women's march} protest:
    \begin{figure}[H]
    \centering
    \includegraphics[width=0.6\textwidth]{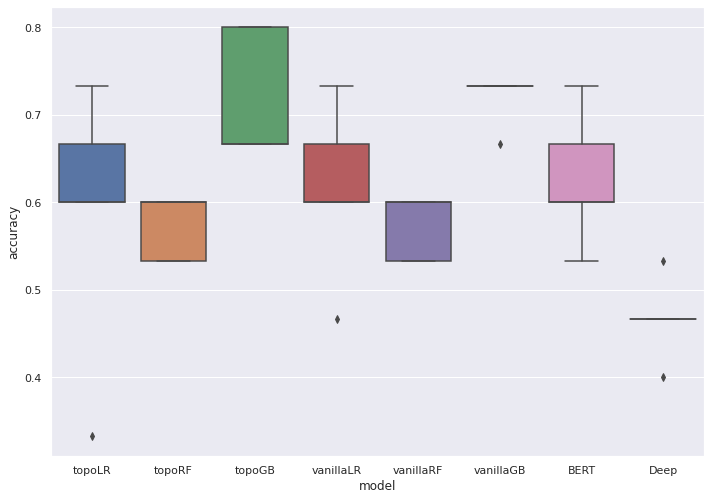}
    \end{figure}

    \item
    1st London Protest:
    \begin{figure}[H]
    \centering
    \includegraphics[width=0.6\textwidth]{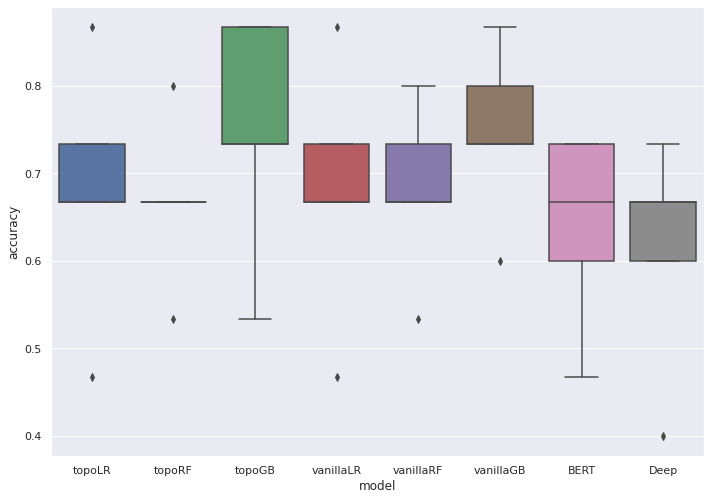}
    \end{figure}

    \item
    2nd London Protest:
    \begin{figure}[H]
    \centering
    \includegraphics[width=0.6\textwidth]{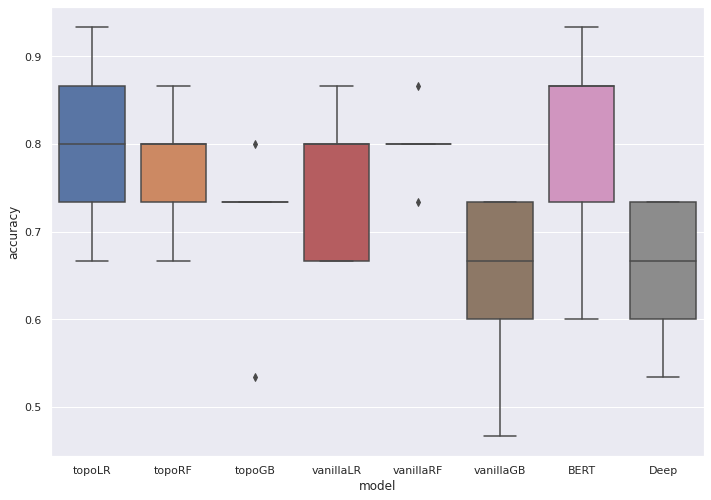}
    \end{figure}
    
    \item
    Gun Control Protest:
    \begin{figure}[H]
    \centering
    \includegraphics[width=0.6\textwidth]{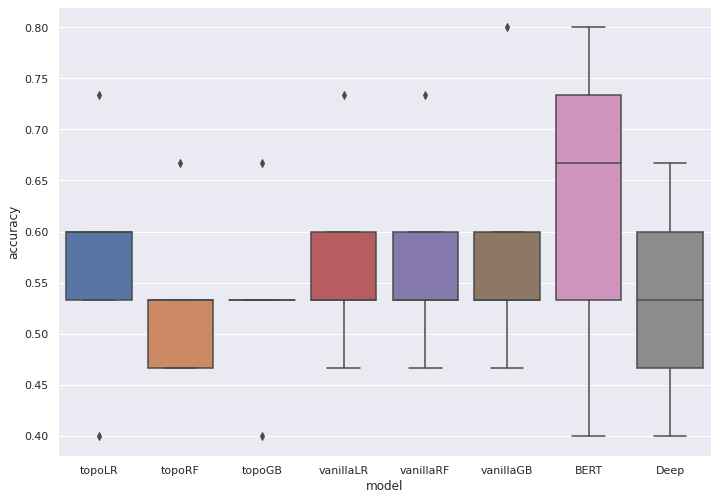}
    \end{figure}

\end{itemize}

\subsection{Low data regimes - out-of-sample}
\label{subsection:Low-Out}
In this section we show the results of the models trained on the \texttt{George Floyd Protests} and validated in the events mentioned below.
\begin{itemize}
    
    \item 
    \texttt{2018 Women's march}:
\begin{table}[H]
\parbox{.3\linewidth}{
\centering
\resizebox{.3\textwidth}{!}{%
\begin{tabular}{ccccc}
\hline
 & precision  &  recall & f1-score  & support \\
\hline
           0   &    0.60   &   0.69   &   0.64    &    26 \\
           1    &   0.73   &   0.65   &   0.69   &     34 \\
    accuracy   &           &          &     0.67        60\\
   macro avg   &    0.67    &  0.67   &   0.67  &      60 \\
weighted avg   &    0.68   &   0.67   &   0.67   &     60 \\
\hline
\end{tabular}
}
\caption{Topological RF} 
}
\hfill
\parbox{.3\linewidth}{
\centering
\resizebox{.3\textwidth}{!}{%
\begin{tabular}{ccccc}
\hline
 & precision  &  recall & f1-score  & support \\
\hline
           0   &      0.70  &    0.70   &   0.70  &      30 \\
           1    &  0.70  &    0.70    &  0.70      &  30\\
    accuracy   &           &          &     0.70    &    60 \\
   macro avg   &  0.70  &    0.70  &    0.70   &     60\\
weighted avg   &    0.70  &    0.70   &   0.70    &    60 \\
\hline
\end{tabular}
}
\caption{Topological GB}
}
\hfill
\parbox{.3\linewidth}{
\centering
\resizebox{.3\textwidth}{!}{%
\begin{tabular}{ccccc}
\hline
 & precision  &  recall & f1-score  & support \\
\hline
           0   &     0.73     &   0.71    &    0.72       &   31 \\
           1    &   0.70      &  0.72     &   0.71     &     29 \\
    accuracy   &           &          &  0.72     &     60 \\
   macro avg   &   0.72     &   0.72     &   0.72     &     60 \\
weighted avg   &    0.72    &    0.72    &    0.72      &    60 \\
\hline
\end{tabular}
}
\caption{Topological LR}
}
\end{table}

\begin{table}[H]
\parbox{.3\linewidth}{
\centering
\resizebox{.3\textwidth}{!}{%
\begin{tabular}{ccccc}
\hline
 & precision  &  recall & f1-score  & support \\
\hline
           0   &    0.60  &    0.69  &    0.64   &     26\\
           1    &  0.73 &     0.65  &    0.69    &    34 \\
    accuracy   &           &          &  0.67   &     60\\
   macro avg   &     0.67    &  0.67  &    0.67   &     60 \\
weighted avg   &  0.68   &   0.67   &   0.67     &   60\\
\hline
\end{tabular}
}
\caption{Vanilla RF}
}
\hfill
\parbox{.3\linewidth}{
\centering
\resizebox{.3\textwidth}{!}{%
\begin{tabular}{ccccc}
\hline
 & precision  &  recall & f1-score  & support \\
\hline
           0   &    0.70    &    0.62   &     0.66       &   34 \\
           1    &   0.57   &     0.65     &   0.61     &     26 \\
    accuracy   &           &          &   0.63    &      60 \\
   macro avg   &   0.63    &    0.64   &     0.63    &      60 \\
weighted avg   &  0.64    &    0.63   &     0.63      &    60 \\
\hline
\end{tabular}
}
\caption{Vanilla GB}
}
\hfill
\parbox{.3\linewidth}{
\centering
\resizebox{.3\textwidth}{!}{%
\begin{tabular}{ccccc}
\hline
 & precision  &  recall & f1-score  & support \\
\hline
           0   &     0.77  &    0.77   &   0.77    &    30\\
           1    &  0.77 &     0.77   &   0.77   &     30\\
    accuracy   &           &          &    0.77    &    60\\
   macro avg   &   0.77  &    0.77   &   0.77   &     60 \\
weighted avg   &   0.77   &   0.77   &   0.77   &     60\\
\hline
\end{tabular}
}
\caption{Vanilla LR}
}
\hfill
\end{table}

\begin{table}[H]
\parbox{.45\linewidth}{
\centering
\resizebox{.45\textwidth}{!}{%
\begin{tabular}{ccccc}
\hline
 & precision  &  recall & f1-score  & support \\
\hline
           0   &   0.53   &   0.73   &   0.62   &     22 \\
           1    &   0.80  &    0.63   &   0.71    &    38 \\
    accuracy   &           &          &    0.67   &     60\\
   macro avg   &     0.67   &   0.68  &    0.66     &   60 \\
weighted avg   &   0.70    &  0.67  &    0.67     &   60 \\
\hline
\end{tabular}
}
\caption{Deep model}
}
\hfill
\parbox{.45\linewidth}{
\centering
\resizebox{.45\textwidth}{!}{%
\begin{tabular}{ccccc}
\hline
 & precision  &  recall & f1-score  & support \\
\hline
           0   &   0.60  &    0.75   &   0.67    &    24 \\
           1    &   0.80   &   0.67   &   0.73    &    36 \\
    accuracy   &           &          &   0.70     &   60\\
   macro avg   &     0.70  &    0.71  &    0.70  &      60 \\
weighted avg   &   0.72   &   0.70  &    0.70    &    60 \\
\hline
\end{tabular}
}
\caption{BERT}
}
\end{table}

        \item \texttt{Put it to the people march}:
\begin{table}[H]
\parbox{.3\linewidth}{
\centering
\resizebox{.3\textwidth}{!}{%
\begin{tabular}{ccccc}
\hline
 & precision  &  recall & f1-score  & support \\
\hline
           0    &   0.53    &   0.84    &   0.65     &    19 \\
           1   &    0.90    &   0.66    &    0.76     &    41 \\
    accuracy   &           &          &  0.72     &    60 \\
   macro avg   &    0.72    &   0.75   &    0.71     &    60 \\
weighted avg   &    0.78    &   0.72    &   0.73      &   60 \\
\hline
\end{tabular}
}
\caption{Topological RF}
}
\hfill
\parbox{.3\linewidth}{
\centering
\resizebox{.3\textwidth}{!}{%
\begin{tabular}{ccccc}
\hline
 & precision  &  recall & f1-score  & support \\
\hline
           0    &    0.63    &   0.79    &   0.70     &    24 \\
           1   &    0.83   &    0.69   &    0.76      &   36 \\
    accuracy   &           &          &   0.73     &    60 \\
   macro avg   &     0.73    &   0.74    &   0.73      &   60\\
weighted avg   &      0.75   &    0.73     &  0.74     &    60 \\
\hline
\end{tabular}
}
\caption{Topological GB}
}
\hfill
\parbox{.3\linewidth}{
\centering
\resizebox{.3\textwidth}{!}{%
\begin{tabular}{ccccc}
\hline
 & precision  &  recall & f1-score  & support \\
\hline
           0    &   0.60  &    0.75   &   0.67    &    24 \\
           1   &    0.80   &   0.67 &     0.73    &    36 \\
    accuracy   &           &          &   0.70     &   60 \\
   macro avg   &  0.70    &  0.71   &   0.70    &    60 \\
weighted avg   &    0.72   &   0.70  &    0.70    &    60 \\
\hline
\end{tabular}
}
\caption{Topological LR}
}
\end{table}

\begin{table}[H]
\parbox{.3\linewidth}{
\centering
\resizebox{.3\textwidth}{!}{%
\begin{tabular}{ccccc}
\hline
 & precision  &  recall & f1-score  & support \\
\hline
           0    &  0.57  &    0.85  &    0.68    &    20 \\
           1   &    0.90   &   0.68   &   0.77    &    40 \\
    accuracy   &           &          &   0.73   &     60 \\
   macro avg   &   0.73  &    0.76 &     0.73   &     60 \\
weighted avg   &    0.79   &   0.73    &  0.74    &    60 \\
\hline
\end{tabular}
}
\caption{Vanilla RF}
}
\hfill
\parbox{.3\linewidth}{
\centering
\resizebox{.3\textwidth}{!}{%
\begin{tabular}{ccccc}
\hline
 & precision  &  recall & f1-score  & support \\
\hline
           0    &   0.63     &  0.66    &   0.64   &      29 \\
           1   &     0.67    &   0.65    &   0.66    &     31 \\
    accuracy   &           &          &  0.65    &     60 \\
   macro avg   &   0.65    &   0.65   &    0.65     &    60 \\
weighted avg   &    0.65   &    0.65   &    0.65    &     60 \\
\hline
\end{tabular}
}
\caption{Vanilla GB}
}
\hfill
\parbox{.3\linewidth}{
\centering
\resizebox{.3\textwidth}{!}{%
\begin{tabular}{ccccc}
\hline
 & precision  &  recall & f1-score  & support \\
\hline
           0    &    0.67   &   0.77   &   0.71 &       26\\
           1   &   0.80  &    0.71  &    0.75     &   34 \\
    accuracy   &           &          &   0.73   &     60 \\
   macro avg   &     0.73   &   0.74  &    0.73    &    60\\
weighted avg   &    0.74   &   0.73  &    0.73     &   60 \\
\hline
\end{tabular}
}
\caption{Vanilla LR}
}
\end{table}

\begin{table}[H]
\parbox{.45\linewidth}{
\centering
\resizebox{.45\textwidth}{!}{%
\begin{tabular}{ccccc}
\hline
 & precision  &  recall & f1-score  & support \\
\hline
           0    &    0.43    &  0.87   &   0.58   &     15 \\
           1   &    0.93   &   0.62  &    0.75     &   45 \\
    accuracy   &           &          &    0.68    &    60 \\
   macro avg   &    0.68    &  0.74  &    0.66   &     60 \\
weighted avg   &    0.81    &  0.68   &   0.70   &     60 \\
\hline
\end{tabular}
}
\caption{Deep model}
}
\hfill
\parbox{.45\linewidth}{
\centering
\resizebox{.45\textwidth}{!}{%
\begin{tabular}{ccccc}
\hline
 & precision  &  recall & f1-score  & support \\
\hline
           0    &  0.63   &    0.76  &     0.69   &      25 \\
           1   &    0.80  &     0.69    &   0.74     &    35 \\
    accuracy   &           &          &   0.72  &       60 \\
   macro avg   &    0.72  &     0.72  &     0.71    &     60\\
weighted avg   &   0.73   &    0.72   &    0.72     &    60 \\
\hline
\end{tabular}
}
\caption{BERT}
}
\end{table}

        \item \texttt{Ukrainian war}:
        \begin{table}[H]
\parbox{.3\linewidth}{
\centering
\resizebox{.3\textwidth}{!}{%
\begin{tabular}{ccccc}
\hline
 & precision  &  recall & f1-score  & support \\
\hline
           0    &   0.47  &    0.93  &    0.62    &    15 \\
           1   &   0.97   &   0.64  &    0.77    &    45 \\
    accuracy   &           &          &   0.72     &   60 \\
   macro avg   &   0.72   &   0.79   &  0.70    &    60 \\
weighted avg   &    0.84  &    0.72  &    0.74    &    60\\
\hline
\end{tabular}
}
\caption{Topological RF}
}
\hfill
\parbox{.3\linewidth}{
\centering
\resizebox{.3\textwidth}{!}{%
\begin{tabular}{ccccc}
\hline
 & precision  &  recall & f1-score  & support \\
\hline
           0    &   0.53   &   0.73   &   0.62    &    22 \\
           1   &     0.80  &    0.63   &   0.71    &    38 \\
    accuracy   &           &          &    0.67   &     60 \\
   macro avg   &    0.67  &    0.68   &   0.66    &    60 \\
weighted avg   &    0.70  &    0.67   &   0.67    &    60 \\
\hline
\end{tabular}
}
\caption{Topological GB}
}
\hfill
\parbox{.3\linewidth}{
\centering
\resizebox{.3\textwidth}{!}{%
\begin{tabular}{ccccc}
\hline
 & precision  &  recall & f1-score  & support \\
\hline
           0    &  0.57   &    0.77   &    0.65     &    22 \\
           1   &    0.83  &     0.66   &    0.74     &    38 \\
    accuracy   &           &          &    0.70   &      60 \\
   macro avg   &   0.70  &     0.72   &    0.69   &      60 \\
weighted avg   &   0.74   &    0.70   &    0.71    &     60 \\
\hline
\end{tabular}
}
\caption{Topological LR}
}
\end{table}

\begin{table}[H]
\parbox{.3\linewidth}{
\centering
\resizebox{.3\textwidth}{!}{%
\begin{tabular}{ccccc}
\hline
 & precision  &  recall & f1-score  & support \\
\hline
           0    &   0.47   &    0.93  &     0.62    &     15 \\
           1   &    0.97  &     0.64   &    0.77    &     45\\
    accuracy   &           &          &    0.72    &     60\\
   macro avg   &    0.72  &     0.79  &     0.70    &     60 \\
weighted avg   &   0.84    &   0.72  &     0.74     &    60 \\
\hline
\end{tabular}
}
\caption{Vanilla RF}
}
\hfill
\parbox{.3\linewidth}{
\centering
\resizebox{.3\textwidth}{!}{%
\begin{tabular}{ccccc}
\hline
 & precision  &  recall & f1-score  & support \\
\hline
           0    &    0.60  &    0.62  &    0.61 &       29 \\
           1   &   0.63   &   0.61   &   0.62     &   31 \\
    accuracy   &           &          &     0.62    &    60 \\
   macro avg   &    0.62   &   0.62   &   0.62      &  60 \\
weighted avg   &    0.62  &    0.62   &   0.62   &     60\\
\hline
\end{tabular}
}
\caption{Vanilla GB}
}
\hfill
\parbox{.3\linewidth}{
\centering
\resizebox{.3\textwidth}{!}{%
\begin{tabular}{ccccc}
\hline
 & precision  &  recall & f1-score  & support \\
\hline
           0    &   0.53   &   0.76  &    0.63   &     21 \\
           1   &  0.83   &   0.64   &   0.72   &     39 \\
    accuracy   &           &          &     0.68   &     60 \\
   macro avg   &    0.68    &  0.70   &   0.68   &     60 \\
weighted avg   &    0.73    &  0.68   &   0.69   &     60 \\
\hline
\end{tabular}
}
\caption{Vanilla LR}
}
\end{table}

\begin{table}[H]
\parbox{.45\linewidth}{
\centering
\resizebox{.45\textwidth}{!}{%
\begin{tabular}{ccccc}
\hline
 & precision  &  recall & f1-score  & support \\
\hline
           0    &  0.47    &  1.00   &   0.64    &    14\\
           1   &    1.00   &   0.65  &    0.79    &    46 \\
    accuracy   &           &          &  0.73   &     60 \\
   macro avg   &    0.73  &    0.83  &    0.71   &     60 \\
weighted avg   &   0.88  &    0.73   &   0.75   &     60\\
\hline
\end{tabular}
}
\caption{Deep model}
}
\hfill
\parbox{.45\linewidth}{
\centering
\resizebox{.45\textwidth}{!}{%
\begin{tabular}{ccccc}
\hline
 & precision  &  recall & f1-score  & support \\
\hline
           0    &   0.50  &    0.68   &   0.58     &   22\\
           1   &     0.77  &    0.61   &   0.68     &   38 \\
    accuracy   &           &          &   0.63     &   60 \\
   macro avg   &    0.63   &   0.64  &    0.63    &    60 \\
weighted avg   &   0.67    &  0.63    &  0.64    &    60 \\
\hline
\end{tabular}
}
\caption{BERT}
}
\end{table}

\end{itemize}

\subsection{High data regimes - in-sample}
\label{subsection:High-In}
In this section we show the results of the models trained and validated in the same events.
\begin{itemize}
\item \texttt{Put it to the people march}:
 \begin{figure}[H]
    \centering
    \includegraphics[width=0.6\textwidth]{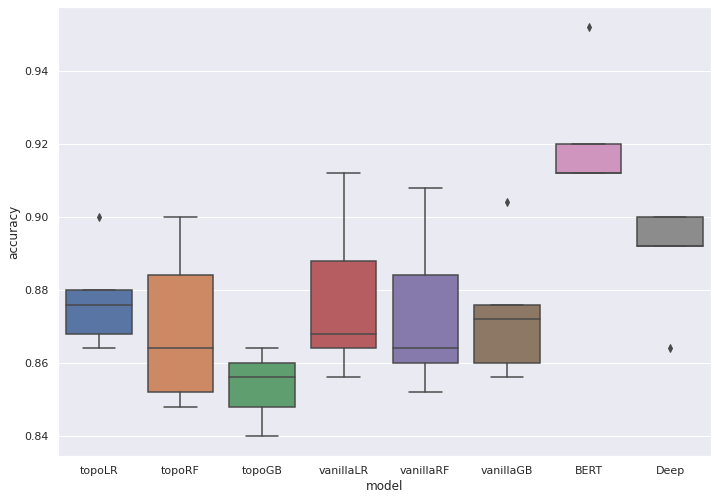}
    \end{figure}
\end{itemize}
\subsection{High data regimes - out-of-sample}
\label{subsection:High-Out}

In this section we show the results of the models trained on the \texttt{George Floyd Protests} and validated in the events mentioned below.
\begin{itemize}
    \item  \texttt{2018 Women's march} protest:
    
    \begin{table}[H]
\parbox{.3\linewidth}{
\centering
\resizebox{.3\textwidth}{!}{%
\begin{tabular}{ccccc}
\hline
 & precision  &  recall & f1-score  & support \\
\hline
           0   &    0.90    &  0.64  &    0.75  &    1409\\
           1    &   0.49  &    0.83 &     0.62    &   591\\
    accuracy   &           &          &   0.70   &   2000 \\
   macro avg   &   0.70   &   0.73    &  0.68   &   2000 \\
weighted avg   &    0.78  &    0.70   &   0.71  &    2000 \\
\hline
\end{tabular}
}
\caption{Topological RF}
}
\hfill
\parbox{.3\linewidth}{
\centering
\resizebox{.3\textwidth}{!}{%
\begin{tabular}{ccccc}
\hline
 & precision  &  recall & f1-score  & support \\
\hline 
           0   &    0.89  &    0.64   &   0.74   &   1396 \\
           1    &   0.49  &    0.81 &     0.61    &   604 \\
    accuracy   &           &          &   0.69  &    2000\\
   macro avg   &    0.69   &   0.72   &   0.68    &  2000 \\
weighted avg   &    0.77    &  0.69   &   0.70   &   2000 \\
\hline
\end{tabular}
}
\caption{Topological GB}
}
\hfill
\parbox{.3\linewidth}{
\centering
\resizebox{.3\textwidth}{!}{%
\begin{tabular}{ccccc}
\hline
 & precision  &  recall & f1-score  & support \\
\hline
           0   &   0.90  &    0.63    &  0.74   &   1431 \\
           1    &   0.47   &   0.82  &    0.60    &   569 \\
    accuracy   &           &          &   0.68  &    2000 \\
   macro avg   &    0.68    &  0.73   &   0.67    &  2000\\
weighted avg   &    0.78   &   0.68   &   0.70   &   2000 \\
\hline
\end{tabular}
}
\caption{Topological LR}
}
\end{table}

\begin{table}[H]
\parbox{.3\linewidth}{
\centering
\resizebox{.3\textwidth}{!}{%
\begin{tabular}{ccccc}
\hline
 & precision  &  recall & f1-score  & support \\
\hline
           0   &    0.90   &   0.64   &   0.75   &   1403 \\
           1    &   0.50   &   0.83  &    0.62    &   597 \\
    accuracy   &           &          &   0.70  &    2000 \\
   macro avg   &    0.70   &   0.74  &    0.68    &  2000\\
weighted avg   &    0.78   &   0.70   &   0.71    &  2000 \\
\hline
\end{tabular}
}
\caption{Vanilla RF}
}
\hfill
\parbox{.3\linewidth}{
\centering
\resizebox{.3\textwidth}{!}{%
\begin{tabular}{ccccc}
\hline
 & precision  &  recall & f1-score  & support \\
\hline
           0   &   0.89   &     0.63  &      0.73  &      1414 \\
           1    &   0.47    &    0.81   &     0.60   &      586 \\
    accuracy   &           &          &   0.68    &    2000 \\
   macro avg   &    0.68    &    0.72     &   0.67    &    2000 \\
weighted avg   &    0.77    &    0.68  &      0.69    &    2000 \\
\hline
\end{tabular}
}
\caption{Vanilla GB}
}
\hfill
\parbox{.3\linewidth}{
\centering
\resizebox{.3\textwidth}{!}{%
\begin{tabular}{ccccc}
\hline
 & precision  &  recall & f1-score  & support \\
\hline
           0   &    0.87  &    0.61   &   0.71  &    1424 \\
           1    &   0.44  &    0.77  &    0.56    &   576 \\
    accuracy   &           &          &   0.65  &    2000 \\
   macro avg   &    0.65   &   0.69  &    0.64   &   2000 \\
weighted avg   &   0.74    &  0.65  &    0.67   &   2000\\
\hline
\end{tabular}
}
\caption{Vanilla LR}
}
\end{table}


\begin{table}[H]
\parbox{.45\linewidth}{
\centering
\resizebox{.45\textwidth}{!}{%
\begin{tabular}{ccccc}
\hline
 & precision  &  recall & f1-score  & support \\
\hline
           0    &   0.89   &   0.62   &   0.73    &  1450 \\
           1   &    0.44   &   0.81   &   0.57    &   550 \\
    accuracy   &           &          &   0.67    &  2000 \\
   macro avg   &    0.67  &    0.71   &   0.65    &  2000\\
weighted avg   &   0.77   &   0.67    &  0.69    &  2000 \\
\hline
\end{tabular}
}
\caption{Deep model}
}
\hfill
\parbox{.45\linewidth}{
\centering
\resizebox{.45\textwidth}{!}{%
\begin{tabular}{ccccc}
\hline
 & precision  &  recall & f1-score  & support \\
\hline
           0    &   0.96   &   0.62 &     0.75    &  1562 \\
           1   &    0.40   &   0.92    &  0.56    &   438 \\
    accuracy   &           &          &  0.68  &    2000 \\
   macro avg   &    0.68    &  0.77  &    0.66  &    2000 \\
weighted avg   &   0.84    &  0.68   &   0.71   &   2000\\
\hline
\end{tabular}
}
\caption{BERT}
}
\end{table}

    \item \texttt{Put it to the people march}
    \begin{table}[H]
\parbox{.3\linewidth}{
\centering
\resizebox{.3\textwidth}{!}{%
\begin{tabular}{ccccc}
\hline
 & precision  &  recall & f1-score  & support \\
\hline
           0   &    0.87  &    0.61   &   0.72   &   1434 \\
           1    &    0.44  &    0.78  &    0.56    &   566\\
    accuracy   &           &          &   0.66   &   2000 \\
   macro avg   &    0.66  &    0.69   &   0.64   &   2000 \\
weighted avg   &   0.75   &   0.66   &   0.67  &    2000\\
\hline
\end{tabular}
}
\caption{Topological RF}
}
\hfill
\parbox{.3\linewidth}{
\centering
\resizebox{.3\textwidth}{!}{%
\begin{tabular}{ccccc}
\hline
 & precision  &  recall & f1-score  & support \\
\hline 
           0   &    0.87   &   0.59 &     0.70  &    1478 \\
           1    &   0.39   &   0.75    &  0.52    &   522 \\
    accuracy   &           &          &   0.63   &   2000 \\
   macro avg   &     0.63   &   0.67 &     0.61    &  2000\\
weighted avg   &     0.75   &   0.63  &    0.65   &   2000 \\
\hline
\end{tabular}
}
\caption{Topological GB}
}
\hfill
\parbox{.3\linewidth}{
\centering
\resizebox{.3\textwidth}{!}{%
\begin{tabular}{ccccc}
\hline
 & precision  &  recall & f1-score  & support \\
\hline
           0   &    0.91  &    0.57   &   0.70   &   1592 \\
           1    &   0.32   &   0.77  &    0.45    &   408 \\
    accuracy   &           &          &   0.61  &    2000 \\
   macro avg   &   0.61  &    0.67  &    0.57   &   2000 \\
weighted avg   &    0.79   &   0.61  &    0.65  &    2000\\
\hline
\end{tabular}
}
\caption{Topological LR}
}
\end{table}

\begin{table}[H]
\parbox{.3\linewidth}{
\centering
\resizebox{.3\textwidth}{!}{%
\begin{tabular}{ccccc}
\hline
 & precision  &  recall & f1-score  & support \\
\hline
           0   &    0.87   &   0.61  &    0.72    &  1425 \\
           1    &    0.45  &    0.78 &     0.57    &   575 \\
    accuracy   &           &          &   0.66  &    2000 \\
   macro avg   &    0.66   &   0.70    &  0.64    &  2000 \\
weighted avg   &    0.75  &    0.66   &   0.68   &   2000 \\
\hline
\end{tabular}
}
\caption{Vanilla RF}
}
\hfill
\parbox{.3\linewidth}{
\centering
\resizebox{.3\textwidth}{!}{%
\begin{tabular}{ccccc}
\hline
 & precision  &  recall & f1-score  & support \\
\hline
           0   &    0.86  &    0.58    &  0.70    &  1477 \\
           1    &   0.39   &   0.74   &   0.51   &    523\\
    accuracy   &           &          &   0.63   &   2000 \\
   macro avg   &    0.63   &   0.66   &   0.60    &  2000 \\
weighted avg   &    0.74   &   0.63    &  0.65    &  2000\\
\hline
\end{tabular}
}
\caption{Vanilla GB}
}
\hfill
\parbox{.3\linewidth}{
\centering
\resizebox{.3\textwidth}{!}{%
\begin{tabular}{ccccc}
\hline
 & precision  &  recall & f1-score  & support \\
\hline
           0   &    0.88    &    0.59    &    0.70     &   1508 \\
           1    &   0.38    &    0.76   &     0.50    &     492 \\
    accuracy   &           &          &   0.63    &    2000 \\
   macro avg   &    0.63    &    0.68    &    0.60     &   2000\\
weighted avg   &   0.76    &    0.63    &    0.66   &     2000 \\
\hline
\end{tabular}
}
\caption{Vanilla LR}
}
\end{table}


\begin{table}[H]
\parbox{.45\linewidth}{
\centering
\resizebox{.45\textwidth}{!}{%
\begin{tabular}{ccccc}
\hline
 & precision  &  recall & f1-score  & support \\
\hline
           0    &   0.89   &   0.57    &  0.69   &   1570 \\
           1   &    0.32  &    0.75   &   0.45    &   430 \\
    accuracy   &           &          &   0.61    &  2000 \\
   macro avg   &    0.61   &   0.66   &   0.57    &  2000\\
weighted avg   &    0.77   &   0.61  &    0.64  &    2000\\
\hline
\end{tabular}
}
\caption{Deep model}
}
\hfill
\parbox{.45\linewidth}{
\centering
\resizebox{.45\textwidth}{!}{%
\begin{tabular}{ccccc}
\hline
 & precision  &  recall & f1-score  & support \\
\hline
           0    &  0.95   &   0.55   &   0.70  &    1723 \\
           1   &    0.23  &    0.82   &   0.36    &   277 \\
    accuracy   &           &          &  0.59    &  2000\\
   macro avg   &    0.59   &   0.69  &    0.53   &   2000 \\
weighted avg   &    0.85   &   0.59  &    0.65   &   2000 \\
\hline
\end{tabular}
}
\caption{BERT}
}
\end{table}


\end{itemize}

\subsection{Artificial time series}
\label{subsection:artif-time-series}
\begin{figure}[H]
\centering
\includegraphics[width=0.86\textwidth]{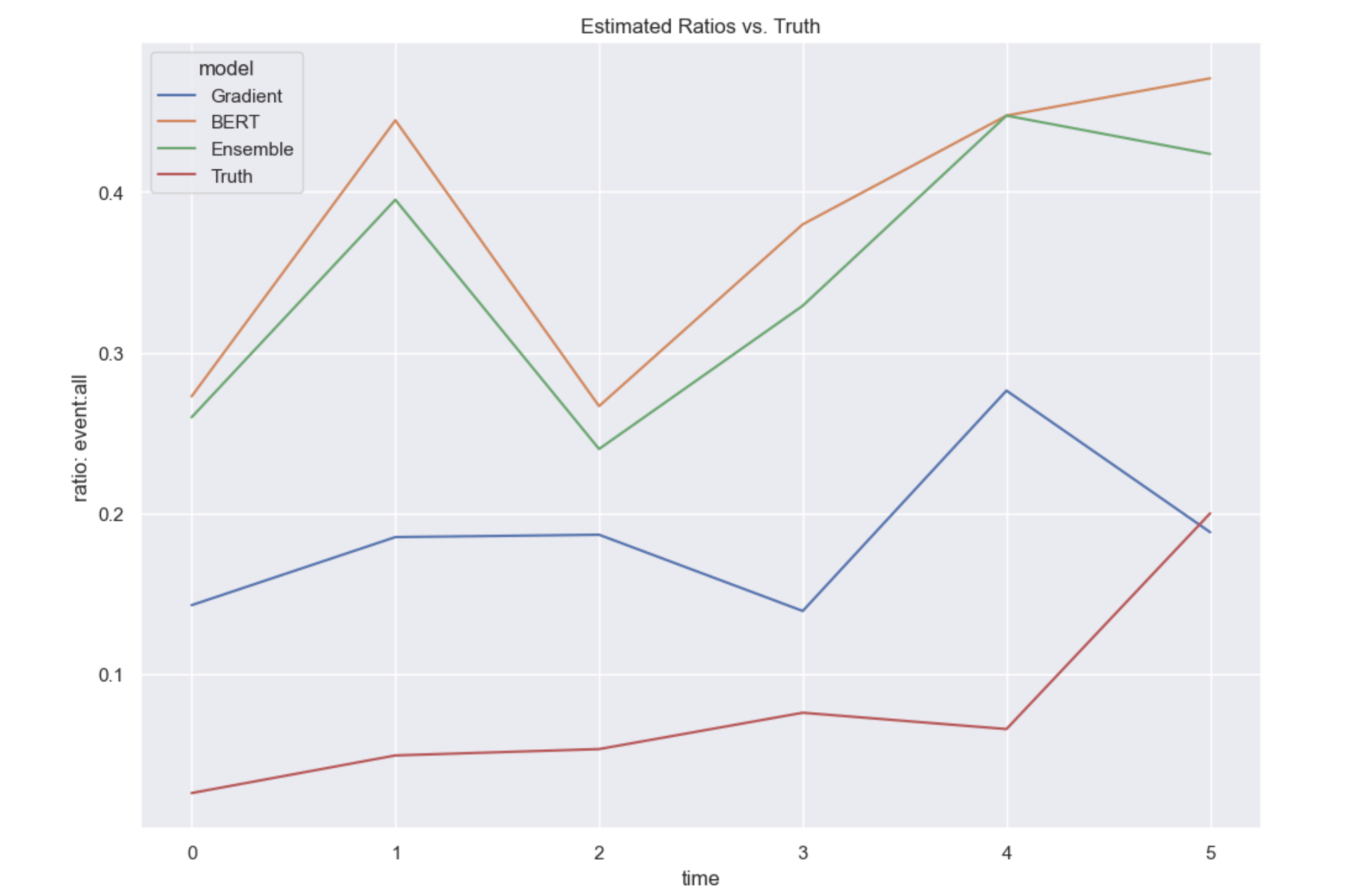}
\includegraphics[width=0.4\textwidth]{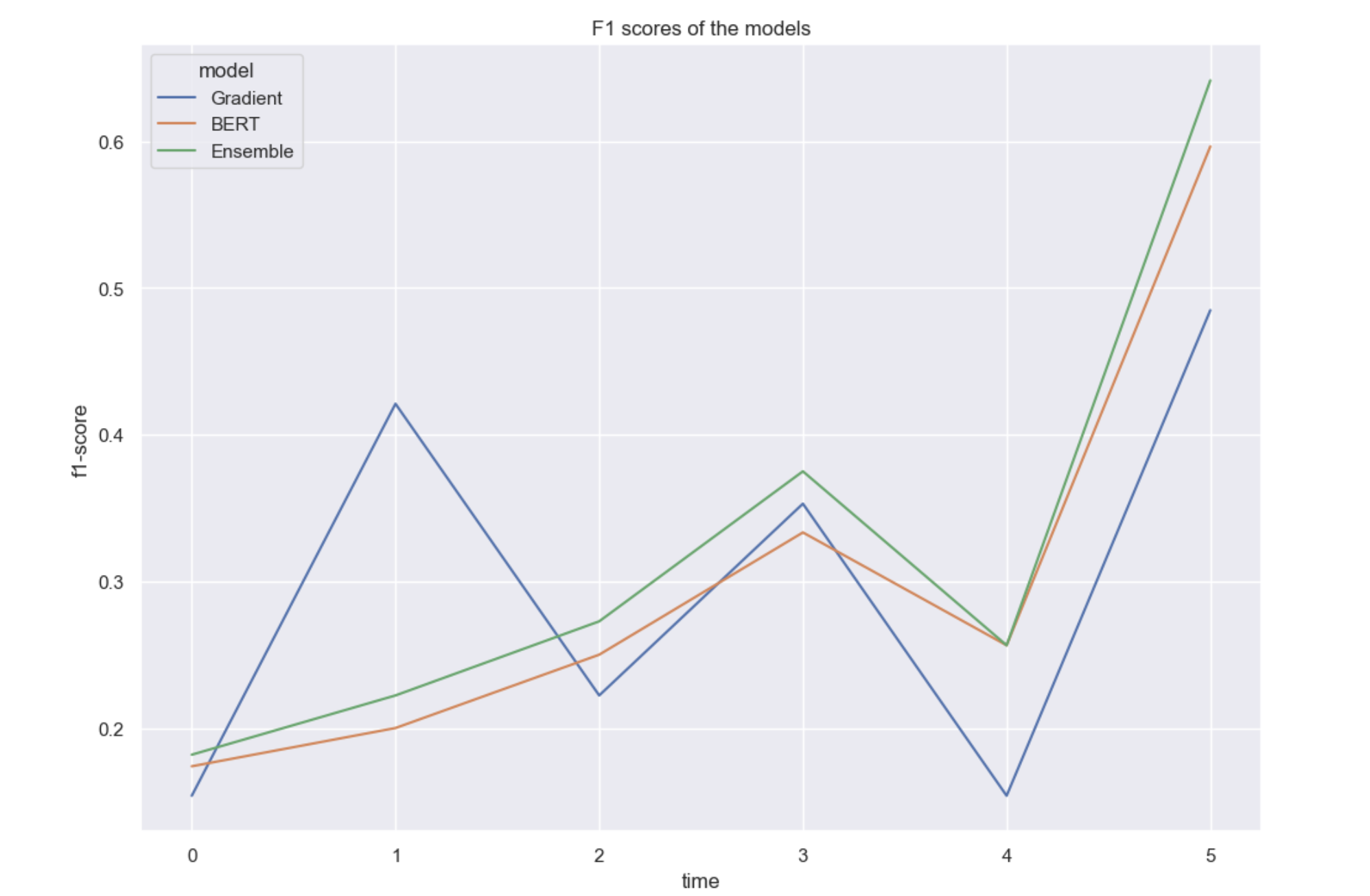} \ \ \ 
\includegraphics[width=0.4\textwidth]{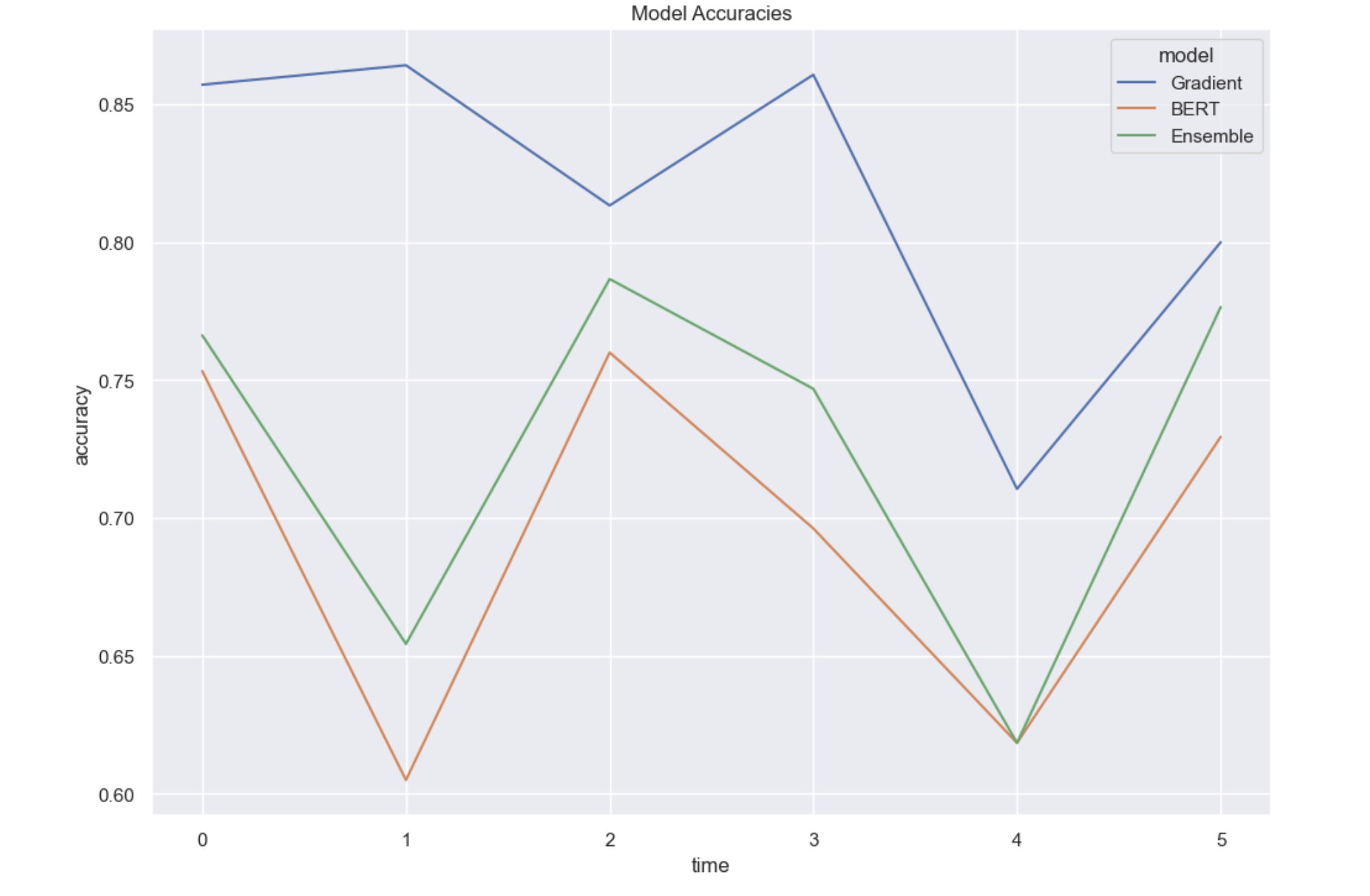}
\caption{\label{fig4}Performance of our models on the \texttt{March for Our Lives} artificial time series with only very few samples per time points ($\sim 10$ samples).}
\end{figure}

As we anticipated, our method outperformed all other models on an artificial time series: another example of an artificial time series can be found in Figure \ref{fig4}.

\end{document}